\RequirePackage{fix-cm}
\documentclass[twocolumn]{svjour3}          
\smartqed  
\usepackage{graphicx}
\usepackage{array}

\usepackage{amsmath, amssymb, dsfont}
\usepackage{color, colortbl}
\definecolor{myColor}{rgb}{0.95,0.95,0.95}

\usepackage{wrapfig,lipsum,booktabs}
\usepackage{epsfig}
\usepackage{floatrow}
\usepackage{array}
\newcolumntype{C}[1]{>{\centering\arraybackslash}m{#1}}
\newcolumntype{L}[1]{>{\arraybackslash}m{#1}}

\usepackage{multirow}

\usepackage[tableposition=top]{caption}%
\usepackage{array,arydshln}
\usepackage{url}
\usepackage{dsfont}
\usepackage{amsfonts}
\usepackage{booktabs}
\usepackage{siunitx}
\usepackage{flushend}

\usepackage{subfigure}
\usepackage{xspace}
\usepackage{tikz}
\usepackage{pgfplots}
\usepackage{times}
\usepackage{marvosym}
\usepackage{microtype}

\usepackage[round]{natbib}

\newcommand{\parbf}[1]{\noindent {\it#1}}
\newcommand{\rev}[1]{#1}

\DeclareMathOperator*{\argmax}{argmax}
\DeclareMathOperator*{\argmin}{argmin}

\usepackage[final, colorlinks=true,allcolors=blue,breaklinks=true]{hyperref} 

\begin{document}

\title{Do semantic parts emerge in Convolutional Neural Networks?
}


\author{Abel Gonzalez-Garcia\and
        Davide Modolo\and
        Vittorio Ferrari
}


\institute{University of Edinburgh, IPAB, School of Informatics \\
              Crichton street 10, Edinburgh EH8 9AB, UK\\
              \email{\{a.gonzalez-garcia,d.modolo\}@sms.ed.ac.uk \\ vittorio.ferrari@ed.ac.uk}
}

\date{Received: date / Accepted: date}

\maketitle

\begin{abstract}
Semantic object parts can be useful for several visual recognition tasks.
Lately, these tasks have been addressed using Convolutional Neural Networks (CNN), achieving outstanding results.
In this work we study whether CNNs learn semantic parts in their internal representation.
We investigate the responses of convolutional filters and try to associate their stimuli with semantic parts. 
We perform two extensive quantitative \rev{analyses}.
First, we use ground-truth part bounding-boxes from the PASCAL-Part dataset to determine how many of those semantic parts emerge in the CNN.
We explore this emergence for different layers, network depths, and supervision levels.
Second, we collect human judgements in order to study what fraction of all filters systematically fire on any semantic part, even if not annotated in PASCAL-Part.
Moreover, we explore several connections between discriminative power and semantics.
We find out which are the most discriminative filters for object recognition, and analyze whether they respond to semantic parts or to other image patches.
We also investigate the other direction: we determine which semantic parts are the most discriminative and whether they correspond to those parts emerging in the network.
This enables to gain an even deeper understanding of the role of semantic parts in the network.

\end{abstract}

\section{Introduction}
\label{intro}
Semantic parts are object regions interpretable by humans (e.g. wheel, leg) and play a fundamental role in several visual recognition tasks. 
For this reason, semantic part-based models have gained significant attention in the last few years.
The key advantages of exploiting semantic part representations is that parts have lower intra-class variability than whole objects, they deal better with pose variation and their configuration provides useful information about the aspect of the object. 
The most notable examples of works on semantic part models are fine-grained recognition~\citep{lin15cvpr,zhang14eccv,parkhi12cvpr}, object class detection~\citep{chen14cvpr}, articulated pose estimation~\citep{liu14eccv,sun11iccv_art,ukita12cvpr}, and attribute prediction~\citep{zhang13iccv,vedaldi14cvpr,gkioxari15iccv}.

Recently, convolutional neural networks (CNNs) have achieved impressive results on many visual recognition tasks, like image classification~\citep{krizhevsky12nips,simonyan15iclr,szegedy15cvpr}, object detection~\citep{girshick14cvpr,he14eccv,girshick15iccv}, semantic segmentation~\citep{long15cvpr,hariharan15cvpr,caesar15bmvc} and fine-grained recognition~\citep{hariharan15cvpr,lin15cvpr,zhang14eccv}.  
Thanks to these outstanding results, CNN-based representations 
are quickly replacing hand-crafted features, like SIFT~\citep{lowe04ijcv} and HOG~\citep{dalal05cvpr}. 

In this paper we look into these two worlds and address the following question: \emph{``does a CNN learn semantic parts in its internal representation?''}  
In order to answer it, we investigate whether the network's convolutional filters learn to respond to semantic parts of objects.
Some previous works~\citep{zeiler14eccv,simonyan14iclr_w} have suggested that semantic parts do emerge in CNNs, but only based on looking at some filter responses on a few images. 
Here we go a step further and perform two quantitative evaluations that examine the different stimuli of the CNN filters and try to associate them with semantic parts.
First, we take advantage of the available ground-truth part location annotations in the PASCAL-Part dataset~\citep{chen14cvpr} to count how many of the annotated semantic parts emerge in a CNN. Second, we use human judgements to determine what fraction of all filters systematically fire on any semantic part (including parts that might not be annotated in PASCAL-Part).

For the first evaluation we use part ground-truth location annotations in the PASCAL-Part dataset \citep{chen14cvpr} to answer the following question: \emph{``how many semantic parts emerge in CNNs?''}.
As an analysis tool, we turn filters into part detectors based on their responses to stimuli.
If some filters systematically respond to a certain semantic part, their detectors will perform well,
and hence we can conclude that they do represent the semantic part.
Given the difficulty of the task, while building the detectors we assist the filters in several ways.
The actual image region to which a filter responds typically does not accurately cover the extent of a semantic part. We refine this region by a regressor trained to map it to a part's ground-truth bounding-box.
Moreover, as suggested by other works~\citep{simon14accv,simon15iccv,xiao15cvpr}, a single semantic part might emerge as distributed across several filters. 
For this reason, we also consider filter combinations as part detectors, and automatically select the optimal combination of filters for a semantic part using a Genetic Algorithm.
We present an extensive analysis on AlexNet~\citep{krizhevsky12nips} finetuned for object detection~\citep{girshick14cvpr}.
Results show that 34 out of 105 semantic parts emerge.
This is a modest number, despite all favorable conditions we have engineered into the evaluation and all assists we have given to the network. 
This result demystifies the impressions conveyed by~\citep{zeiler14eccv,simonyan14iclr_w} and shows that the network learns to associate filters to part classes, but only for some of them and often to a weak degree.
In general, these semantic parts are those that are large or very discriminative for the object class (e.g., torso, head, wheel).
Finally, we analyze different network layers, architectures, and supervision levels. We observe that part emergence increases with the depth of the layer, especially when using deeper architectures such as VGG16~\citep{simonyan15iclr}. Moreover, emergence decreases when the network is trained for tasks less related to object parts, e.g. scene classification~\citep{zhou14nips}.

Our second quantitative evaluation answers the converse question: \emph{``what fraction of all filters respond to any semantic part?''}. As \rev{PASCAL-Part} is not fully annotated (e.g. car door handle is missing), we answer it using human judgements. 
For each filter, we show human annotators the 10 images with the highest activations per object class.
We highlight the regions corresponding to the activations and ask the annotators whether they systematically cover the same concept (e.g. a semantic part, a background, a texture, a color, etc.). 
In case of positive answer, we ask them to name the concept (e.g. horse hoof). 
In general, the majority of the filters do not seem to systematically respond to any concept.
On average per object class, 7\% of the filters correspond to semantic parts (including several filters responding to the same semantic part). About 10\% of the \rev{filters} systematically respond to other stimuli such as colors, subregions of parts or even assemblies of multiple parts.
Finally, we also compare the semantic parts emerging in this evaluation with the 34 parts annotated in PASCAL-Part that emerged in the first evaluation.
We find that nearly all the parts that emerge according to the detection performance criterion used in the first evaluation also emerge according to human judgements. However, more semantic parts emerge according to human judgements, including several parts that are not annotated in PASCAL-Part.

Finally, we also investigate \rev{how discriminative network filters and semantic parts are for recognizing objects}.
We explore the possibility that some filters respond to `parts' as recurrent discriminative patches, rather than truly semantic parts.
We find that, for each \rev{object} class \rev{in PASCAL-Part}, there are on average 9 discriminative filters that are largely responsible for recognizing it.
Interestingly, 40\% of these are also semantic according to human judgements, which is a much greater proportion than the 7\% found when considering {\em all} filters. The overlap between which filters are discriminative and which ones are semantic might be the reason why previous works~\citep{zeiler14eccv,simonyan14iclr_w} have suggested a stronger emergence of semantic parts, based on qualitative visual inspection. 
We also investigate to what degree the emergence of semantic parts in the network correlates with their discriminativeness for recognition. Interestingly, these are highly correlated: semantic parts that are discriminative emerge much more than other semantic parts. While this is generally assumed in the community, ours is the first work presenting a proper quantitative evaluation that turns this assumption into a fact.

The rest of the paper is organized as follows.
Section~\ref{sec:cnn:relatedWork} discusses some related work.
Section~\ref{sec:cnn:pascalParts} presents our quantitative evaluation using PASCAL-Part bounding-boxes, while
evaluation using human judgements is presented in section~\ref{sec:cnn:humans}.
The discriminativeness of filters is investigated in section~\ref{sec:cnn:discr_filters}, while the discriminativeness of semantic parts in section~\ref{sec:cnn:discrim_parts}.
Finally, section~\ref{sec:cnn:conclusions} summarizes the conclusions of our study.

\section{Related Work}
\label{sec:cnn:relatedWork}

\noindent{\it Analyzing CNNs.} 
CNN-based representations are unintuitive and there is no clear understanding of why they perform so well or how they could be improved. 
In an attempt to better understand the properties of a CNN, some recent vision works have focused on analyzing their internal representations~\citep{szegedy14iclr,yosinski14nips,lenc15cvpr,mahendran15cvpr,zeiler14eccv,simonyan14iclr_w,agrawal14eccv,zhou15iclr,eigen2013iclr_w}. 
Some of these investigated properties of the network, like stability~\citep{szegedy14iclr}, feature transferability~\citep{yosinski14nips}, equivariance, invariance and equivalence~\citep{lenc15cvpr}, the ability to reconstruct the input~\citep{mahendran15cvpr} and how the number of layers, filters and parameters affects the network performance~\citep{agrawal14eccv,eigen2013iclr_w}. 

\cite{zeiler14eccv} use deconvolutional networks to visualize locally optimal visual inputs for individual filters. 
\cite{simonyan14iclr_w} use a gradient-based visualization technique to highlight the areas of an image discriminative for an object class.
\cite{agrawal14eccv} show that the feature representations are distributed across object classes. 
\cite{zhou15iclr} show that the layers of a network learn to recognize visual elements at different levels of abstraction (e.g. edges, textures, objects and scenes).
Most of these works make an interesting observation: filter responses can often be linked to semantic parts~\citep{zeiler14eccv,simonyan14iclr_w,zhou15iclr}. 
These observations are however mostly based on casual visual inspection of few images~\citep{zeiler14eccv,simonyan14iclr_w}. 
\citep{zhou15iclr} is the only work presenting some quantitative results based on human judgements, but not focused on semantic parts. 
 Instead, we present an extensive quantitative analysis on whether filters can be associated with semantic parts and to which degree. We transform the filters into part detectors and evaluate their performance on ground-truth part bounding-boxes from the PASCAL-Part dataset~\citep{chen14cvpr}. Moreover, we present a second quantitative analysis based on human judgements where we categorize filters into semantic parts. 
 We believe this methodology goes a step further than previous works and supports more conclusive answers to the quest for semantic parts.\\

\noindent{\it Filters as intermediate part representations for recognition.} 
Several works use filter responses for recognition tasks \citep{simon14accv,gkioxari15iccv,simon15iccv,xiao15cvpr,oquab15cvpr}. 
\cite{simon14accv} train part detectors for fine-grained recognition, while \cite{gkioxari15iccv} train them for action and attribute classification.
Furthermore, \cite{simon15iccv} learn constellations of filter activation patterns, and \cite{xiao15cvpr} cluster group of filters responding to different bird parts.
All these works assume that the convolutional layers of a network are related to semantic parts.
In this paper we try to shed some light on this assumption and hopefully inspire more works on exploiting the network's internal structure for recognition.
\begin{figure*}
  \begin{center}
    \includegraphics[width=\textwidth]{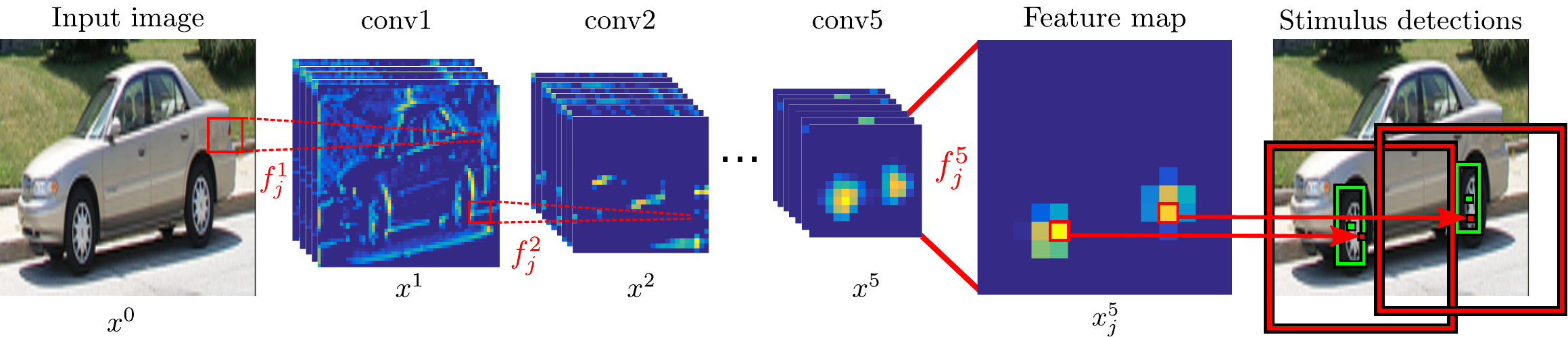}
\end{center}
\caption{\small Overview of our approach for a layer 5 filter. Each local maxima of the filter's feature map leads to a stimulus detection (\textcolor{red}{red}).
  We transform each detection with a regressor trained to map it to a bounding-box tightly covering a semantic part (\textcolor{green}{green}).}
\label{fig:cnn:overview}
\end{figure*}

\section{\rev{PASCAL-Part} emergence in CNNs}
\label{sec:cnn:pascalParts}

Our goal is understanding whether the convolutional filters learned by the network respond to semantic parts.
In order to do so, we investigate the image regions to which a filter responds and try to associate them with a particular part.\\

\parbf{Network architecture.}
Standard image classification CNNs such as~\citep{krizhevsky12nips,simonyan15iclr} process an input image through a sequence of layers of various types, and finally output a class probability vector.
Each layer $i$ takes the output of the previous layer $x^{i-1}$ as input, and produces its output $x^i$ by applying up to four operations: convolution, nonlinearity, pooling, and normalization.
The convolution operation slides a set of learned filters of different sizes and strides over the input.
The nonlinearity of choice for many networks is the Rectified Linear Unit (ReLU)~\citep{krizhevsky12nips}, and it is applied right after the convolution.\\

\subsection{Methodology}
\label{sec:cnn:approach}
Fig.~\ref{fig:cnn:overview} presents an overview of our approach.
Let $f_j^i$ be the $j$-th convolutional filter of the $i$-th layer, including also the ReLU.
Each pixel in a feature map $x_j^i = f_j^i(x^{i-1})$ is the activation value of filter $f_j^i$ applied to a particular position in the feature maps $x^{i-1}$ of the previous layer. The resolution of the feature map depends on the layer, decreasing as we advance through the network.
Fig.~\ref{fig:cnn:overview} shows feature maps for layers 1, 2, and 5.
When a filter responds to a particular stimulus in its input, the corresponding region on the feature map has a high activation value.
By studying the stimuli that cause a filter to fire, we can characterize them and decide whether they correspond to a semantic object part.

\subsubsection{Stimulus detections from activations} 
\label{sec:cnn:reg}
The value $a_{c,r}$ of each particular activation $\alpha$, located at position $(c, r)$ of feature map $x_j^i$, indicates the response of the filter to a corresponding region in its input $x^{i-1}$.
The receptive field of an activation is the region on the input image on which the filter acted, and it is determined by the network structure.
By recursively back-propagating the input region of activation $\alpha$ down the layers, we can reconstruct the actual receptive field on the input image.
The size of the receptive field varies depending on the layer, from the actual size of the filter for the first convolutional layer, up to a much larger image region on the top layer.
For each feature map, we select all its local maxima activations.
Each of these activations will lead to a stimulus detection in the image, regardless of its activation value (i.e. no minimum threshold).
Therefore, all peaks of the feature map become detections, and their detection scores are their activation values.
The location of such detections is defined by the center of the receptive field of the corresponding activation, whereas its size varies depending on the layer.
Fig.~\ref{fig:cnn:overview} shows an example, where the two local maxima of feature map $x^5_j$ lead to the stimulus detections depicted in red, which correspond to their receptive fields on the image.\\

\begin{figure}
  \begin{center}
    \includegraphics[width=\textwidth]{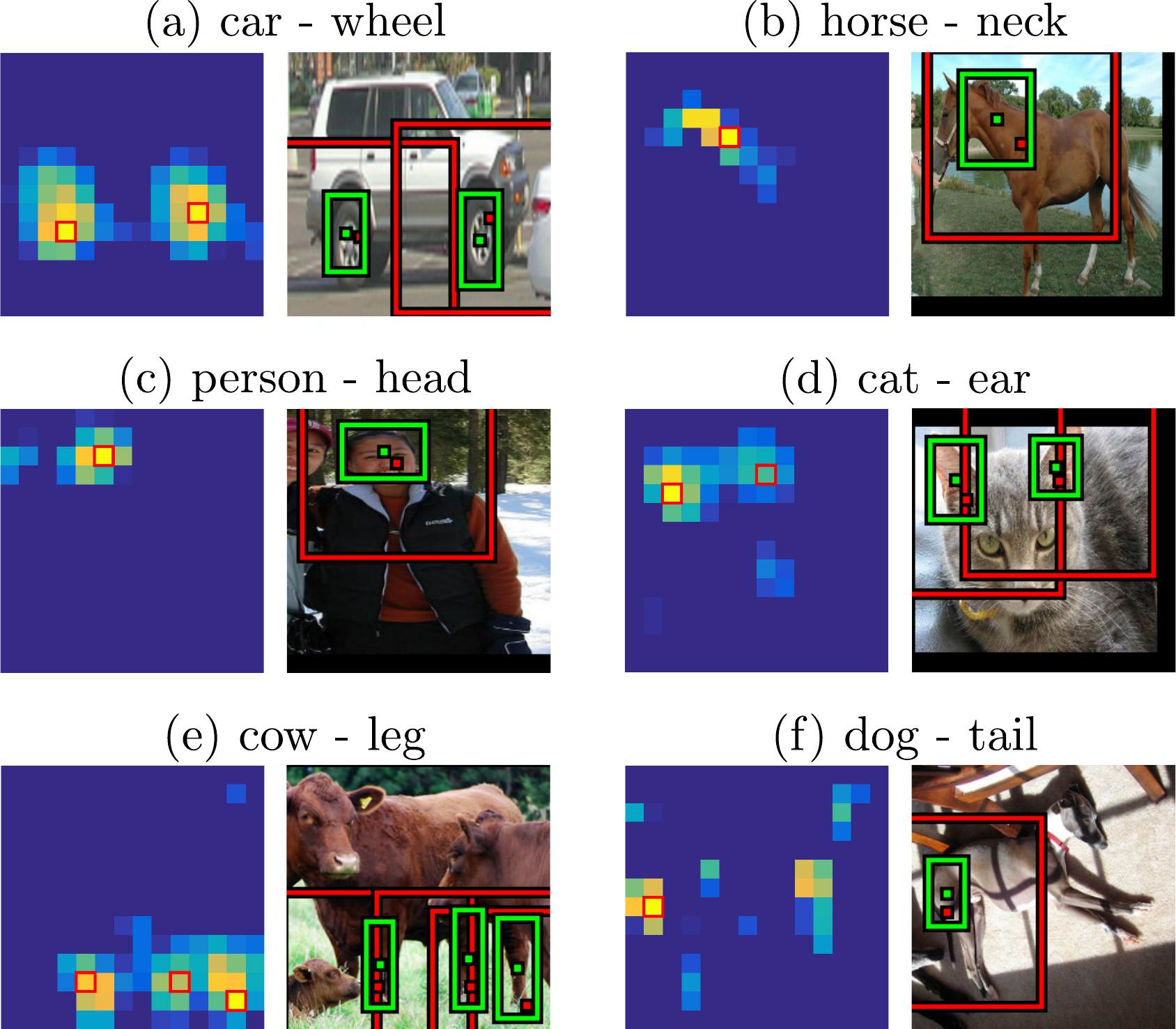}
\end{center}
  \caption{\small Examples of stimulus detections for layer 5 filters. For each part class we show a feature map on the left, where we highlight some local maxima in red. On the right, instead, we show the corresponding original \textcolor{red}{receptive field} and the \textcolor{green}{regressed box}.}
\label{fig:cnn:qualResRegression}
\end{figure}

\noindent{\it Regressing to part bounding-boxes.}
The receptive field of an activation gives a rough indication about the location of the stimulus.
However, it rarely covers a part tightly enough to associate the stimulus with a part instance (fig.~\ref{fig:cnn:qualResRegression}).
In general, the receptive field of high layers is significantly larger than the part ground-truth bounding-box, especially for small classes like ear.
Moreover, while the receptive field is always square, some classes have other aspect ratios (e.g. legs).
Finally, the response of a filter to a part might not occur in its center, but at an offset instead (e.g. on the bottom area, fig.~\ref{fig:cnn:qualResRegression}{\color{blue}d-e}).

In order to factor out these elements, we assist each filter with a bounding-box regression mechanism that refines its stimulus detection for each part class. 
This regressor turns each stimulus detection, which are generally bigger than the corresponding part and have a squared aspect ratio (fig.~\ref{fig:cnn:overview}) into more accurate detections that fit the part instances more tightly (fig.~\ref{fig:cnn:qualResRegression}).
The regressor applies a 4D transformation, i.e. translation and scaling along width and height.
We believe that if a filter fires systematically on many instances of a part class at the same relative location (in 4D), then we can grant that filter a `part detector' status.
This implies that the filter responds to that part, even if the actual receptive field does not tightly cover it.
For the rest of the paper, all stimulus detections include this regression step unless stated otherwise.

We train one regressor for each part class and filter.
Let $\{G^l\}$ be the set of all ground-truth bounding-boxes for the part in the training set.
Each instance bounding-box $G^l$  is defined by its center coordinates $(G^l_x, G^l_y)$, width $G^l_w$, and height $G^l_h$.
We train the regressor on $K$ pairs of activations and ground-truth part bounding-boxes $\{\alpha^k, G^k\}$.
Let $(c_x,c_y)$ be the center of the receptive field on the image for a particular feature map activation $\alpha$ of value $a_{c,r}$, and let $w,h$ be its width and height ($w=h$ as all receptive fields are square). 
We pair each activation with an instance bounding-box $G^l$ of the corresponding image if $(c_x,c_y)$ lies inside it.
We are going to learn a 4D transformation $d_x, d_y, d_w, d_h$ to predict a part bounding-box $G'$ from $\alpha$'s receptive field 

\noindent
\begin{minipage}{.5\linewidth}
\begin{equation*}
  \small
  G_x' = x + d_x(\gamma(\alpha))
\end{equation*}
\end{minipage}%
\begin{minipage}{.5\linewidth}

\begin{equation*}
  \small
    \centering
  G_w' = d_w(\gamma(\alpha)) 
\end{equation*}
\end{minipage}

\noindent\begin{minipage}{.5\linewidth}

\begin{equation*}
  \small
    \centering
  G_y' = y + d_y(\gamma(\alpha)) 
\end{equation*}
\end{minipage}%
\begin{minipage}{.5\linewidth}
\begin{equation*}
  \small
    \centering
  G_h' = d_h(\gamma(\alpha)), 
\end{equation*}
\end{minipage}

\vspace{3mm}
\noindent where $\gamma(\alpha) = (c_x, c_y, a_{c-1,r-1}, a_{c-1,r},..., a_{c+1,r+1})$. Therefore, the regression depends on the center of the receptive field and on the values of the 3x3 neighborhood of the activation on the feature map. 
Note that it is independent of $w$ and $h$ as these are fixed for a given layer.
Each $d_*$ is a linear combination of the elements in $\gamma(\alpha)$ with a weight vector $w_*$, where $*$ can be $x,y,w$, or $h$.

We set regression targets $(t_x^k, t_y^k, t_w^k, t_h^k) = (G_x^k - c_x^k, G_y^k - c_y^k, G_w^k, G_h^k)$ and optimize the following weighted least squares objective
\begin{equation}
  \small
    \centering
  w_* = \argmin_{w_*'}\sum^K_{k=1} a^k_{c,r}(t_*^k - w_*'\cdot \gamma(\alpha^k))^2.
\end{equation}
In practice, this tries to transform the position, size and aspect-ratio of the original receptive field of the activations into the bounding-boxes in $\{G^l\}$.

Fig.~\ref{fig:cnn:qualResRegression} presents some examples of our bounding-box regression for 6 different parts.
For each part, we show the feature map of a layer 5 filter and both the original receptive field (red) and the regressed box (green) of some activations. 
We can see how given a local maximum activation on the feature map, the regressor not only refines the center of the detection, but also successfully captures its extent.
Some classes are naturally more challenging, like \emph{dog-tail} in fig.~\ref{fig:cnn:qualResRegression}{\color{blue}f}, due to higher size and aspect-ratio variance or lack of satisfactory training examples.\\

\parbf{Evaluating filters as part detectors.}
For each filter and part combination, we need to evaluate the performance of the filter as a detector of that part. 
We take all the local maxima of the filter's feature map for every input image and compute their stimulus detections, applying Non-Maxima Suppression~\citep{felzenszwalb10pami} to remove duplicate detections.
We consider a stimulus detection as correct if it has an intersection-over-union $\geq 0.4$ with any ground-truth bounding-box of the part, following~\cite{chen14cvpr}.
All other detections are considered false positives\rev{.}
A filter is a good part detector if it has high recall but a small number of false positives, indicating that when it fires, it is because the part is present.
Therefore, we use Average Precision (AP) to evaluate the filters as part detectors, following the PASCAL VOC~\citep{everingham10ijcv} protocol.

\subsubsection{Filter combinations}
\label{sec:cnn:GA}
Several works \citep{agrawal14eccv,zhou15iclr,xiao15cvpr} noted that one filter alone is often insufficient to cover the spectrum of appearance variation of an object class. We believe that this holds also for part classes.
For this reason, we present here a technique to automatically select the optimal combination of filters for a part class. 

For a given network layer, the search space consists of binary vectors $\mathbf{z} = [z_1, z_2, ..., z_N]$, where $N$ is the number of filters in the layer.
If $z_i=1$, then the $i$-th filter is included in the combination.
We consider the stimulus detections of a filter combination as the set union of the individual detections of each filter in it.
Ideally, a good filter combination should make a better part detector than the individual filters in it.
Good combinations should include complementary filters that jointly detect a greater number of part instances, increasing recall. 
At the same time, the filters in the combination should not add many false positives.
Therefore, we can use the collective AP of the filter combination as objective function to be maximized:
\vspace{-2mm}
\begin{equation}
  \small
  \mathbf{z} = \argmax_{\mathbf{z'}} \text{AP}(\bigcup_{i\in \{j|z'_j =1\}}\text{det}_i),
  \vspace{-3mm}
  \label{eq:collectiveAP}
\end{equation}
where $\text{det}_i$ indicates the stimulus detections of the $i$-th filter.

We use a Genetic Algorithm (GA)~\citep{mitchell98genetic} to optimize this objective function.
GAs are iterative search methods inspired by natural evolution. 
At every generation, the algorithm evaluates the `fitness' of a set of search points (population). 
Then, the GA performs three genetic operations to create the next generation: selection, crossover and mutation. 
In our case, each member of the population (chromosome) is a binary vector $\mathbf{z}$ as defined above. 
Our fitness function is the AP of the filter combination.
In our experiments, we use a population of 200 chromosomes and run the GA for 100 generations.
We use Stochastic Universal Sampling~\citep{mitchell98genetic}.
We set the crossover and mutation probabilities to 0.7 and 0.3, respectively.
We bias the initialization towards a small number of filters by setting the probability $P(z_i=1) = 0.02, \forall i$. This leads to an average combination of 5 filters when $N=256$, in the initial population.

\rev{We underline that our goal is to find filter combinations that act collectively as part detectors, which is formalized in the objective~\eqref{eq:collectiveAP}. While a GA is a suitable method to maximize~\eqref{eq:collectiveAP}, other methods could be used instead.}

\subsection{AlexNet for object detection}
\label{sec:cnn:alexNetDet}

In this section we analyze the role of convolutional filters in AlexNet and test whether some of them can be associated with semantic parts.
In order to do so, we design our settings to favor the emergence of this association.\\

\begin{table*}
\centering
{
\footnotesize
\resizebox{\columnwidth}{!}{
  \begin{tabular}{ll:ccc:ccc:ccc:ccc:ccc}
    \hline
\multirow{2}{*}{Class} & \multirow{2}{*}{Part}& \multicolumn{3}{c:}{Layer 1 (96)}& \multicolumn{3}{c:}{Layer 2 (256)} & \multicolumn{3}{c:}{Layer 3 (384)} & \multicolumn{3}{c:}{Layer 4 (384)} & \multicolumn{3}{c}{Layer 5 (256)}   \\ 
& & Best & GA & nFilters & Best & GA & nFilters & Best & GA & nFilters & Best & GA & nFilters & Best & GA & nFilters \\
\hline
\multirow{4}{*}{aero} 
& body & 17.7 & {\tiny +}3.4 & 12 & 23.7 & {\tiny +}10.2 & 33 & 29.4 & {\tiny +}9.5 & 62 & 34.0 & {\tiny +}9.2 & 49 & 29.3 & {\tiny +}17.0 & 49 \\ 
& stern & 10 & {\tiny +}1.5 & 14 & 13.6 & {\tiny +}5.1 & 33 & 21.4 & {\tiny +}2.5 & 45 & 19.2 & {\tiny +}5.2 & 32 & 15.0 & {\tiny +}9.4 & 21 \\ 
& wing & 4.2 & {\tiny +}1.2 & 12 & 5.6 &{\tiny +}5.3 & 41 & 6.0 & {\tiny +}7.0 & 63 & 6.9 & {\tiny +}3.9 & 36 & 4.7 & {\tiny +}9.6 & 38 \\
& engine & 2.1 & {\tiny +}0.9 & 14 & 2.7 & {\tiny +}1.6 & 5 & 4.2 & {\tiny +}1.7 & 37 & 4.5 & {\tiny +}2.5 & 53 & 1.6 & {\tiny +}5.4 & 25 \\ 
\hline
\multirow{4}{*}{bike} 
& wheel & 14.2 & {\tiny +}0.7 & 19 & 41.4 & {\tiny +}0.0 & 30 & 49.2 & {\tiny +}0.0 & 3 & 60.0 & {\tiny +}0.0 & 10 & 57.1 & {\tiny +}6.1 & 16 \\ 
& saddle & 1.0 & {\tiny +}0.5 & 5 & 1.7 & {\tiny +}0.5 & 18 & 1.6 & {\tiny +}0.6 & 43 & 1.7 & {\tiny +}0.0 & 4 & 2.1 & {\tiny +}2.5 & 16 \\ 
& handlebar & 2.8 & {\tiny +}0.4 & 17 & 3 & {\tiny +}2.2 & 21 & 4.0 & {\tiny +}2.0 & 37 & 3.2 & {\tiny +}3.1 & 40 & 4.1 & {\tiny +}5.8 & 38 \\
& chainwheel & 0.6 & {\tiny +}0.1 & 4 & 0.6 & {\tiny +}0.6 & 24 & 1.9 & {\tiny +}0.0 & 46 & 1.7 & {\tiny +}1.0 & 17 & 3.0 & {\tiny +}0.6 & 6 \\
\hline
\multirow{2}{*}{bottle} 
& cap & 1.8 & {\tiny +}0.6 & 13 & 4.4 & {\tiny +}2.2 & 20 & 6.4 & {\tiny +}0.9 & 21 & 11.2 & {\tiny +}0.0 & 11 & 6.6 & {\tiny +}4.6 & 15 \\ 
& body & 73.0 & {\tiny +}0.6 & 4 & 80.9 & {\tiny +}0.0 & 9 & 87.6 & {\tiny +}0.0 & 10 & 83.4 & {\tiny +}0.0 & 3 & 81.0 & {\tiny +}6.3 & 25 \\ 
\hline
\multirow{4}{*}{cat} 
& head & 16.8 & {\tiny +}0.0 & 1 & 21.2 & {\tiny +}0.0 & 8 & 30.6 & {\tiny +}6.0 & 8 & 44.5 & {\tiny +}1.1 & 15 & 53.9 & {\tiny +}5.2 & 10 \\
& eye & 0.6 & {\tiny +}0.0 & 4 & 10.4 & {\tiny +}1.6 & 3 & 10.8 & {\tiny +}0.0 & 2 & 3.8 & {\tiny +}0.5 & 18 & 4.3 & {\tiny +}1.3 & 4 \\
& ear & 1.9 & {\tiny +}0.6 & 10 & 4.4 & {\tiny +}0.3 & 14 & 4.9 & {\tiny +}5.7 & 12 & 10.7 & {\tiny +}5.1 & 13 & 17.5 & {\tiny +}2.8 & 10 \\
& paw & 0.6 & {\tiny +}0.1 & 6 & 0.7 & {\tiny +}0.4 & 18 & 1.9 & {\tiny +}1.2 & 21 & 3.2 & {\tiny +}0.0 & 11 & 1.5 & {\tiny +}1.9 & 16 \\
& tail & 1.0 & {\tiny +}0.0 & 1 & 1.3 & {\tiny +}1.4 & 35 & 2.1 & {\tiny +}2.8 & 71 & 2.3 & {\tiny +}4.0 & 42 & 2.2 & {\tiny +}3.9 & 24 \\
\hline
\multirow{4}{*}{cow} 
& head & 12.9 & {\tiny +}0.4 & 12 & 15.8 & {\tiny +}3.8 & 34 & 21.2 & {\tiny +}8.5 & 71 & 22.9 & {\tiny +}4.9 & 50 & 24.6 & {\tiny +}17.1 & 34 \\
& muzzle & 3.4 & {\tiny +}0.2 & 14 & 4.9 & {\tiny +}2.9 & 37 & 15.4 & {\tiny +}0.0 & 10 & 15.6 & {\tiny +}1.9 & 14 & 16.7 & {\tiny +}9.5 & 22 \\ 
& torso & 43.1 & {\tiny +}0.0 & 1 & 56.6 & {\tiny +}0.0 & 45 & 62.0 & {\tiny +}9.0 & 42 & 63.6 & {\tiny +}9.9 & 53 & 65.2 & {\tiny +}13.6 & 42 \\ 
& tail & 0.9 & {\tiny +}0.0 & 9 & 2.9 & {\tiny +}1.1 & 19 & 2.9 & {\tiny +}2.2 & 16 & 7.0 & {\tiny +}2.5 & 30 & 3.7 & {\tiny +}0.8 & 6 \\ 
\hline
\multirow{4}{*}{horse} 
& head & 5.4 & {\tiny +}0.3 & 3 & 7.6 & {\tiny +}1.8 & 22 & 10.7 & {\tiny +}3.1 & 52 & 15.3 & {\tiny +}5.2 & 27 & 16.1 & {\tiny +}11.6 & 22 \\
& ear & 0.9 & {\tiny +}0.3 & 11 & 1.3 & {\tiny +}1.1 & 18 & 4.3 & {\tiny +}1.6 & 9 & 2.7 & {\tiny +}3.2 & 12 & 6.1 & {\tiny +}0.5 & 4 \\
& torso & 48.7 & {\tiny +}0.9 & 9 & 52.7 & {\tiny +}4.1 & 22 & 63.8 & {\tiny +}0.0 & 11 & 63.0 & {\tiny +}4.4 & 27 & 65.2 & {\tiny +}7.1 & 29 \\
& leg & 6.3 & {\tiny +}0.2 & 10 & 10.7 & {\tiny +}3.6 & 6 & 14.2 & {\tiny +}7.5 & 8 & 23.0 & {\tiny +}5.7 & 9 & 23.4 & {\tiny +}9.6 & 14 \\
\hline

\multirow{4}{*}{person} 
& head & 6.6 & {\tiny +}0.0 & 1 & 8.7 & {\tiny +}0.0 & 3 & 33.8 & {\tiny +}0.0 & 5 & 44.9 & {\tiny +}0.0 & 6 & 58.2 & {\tiny +}0.0 & 1 \\
& hair & 3.9 & {\tiny +}0.1 & 4 & 5.1 & {\tiny +}0.0 & 3 & 18.0 & {\tiny +}0.0 & 11 & 28.7 & {\tiny +}0.0 & 4 & 30.6 & {\tiny +}0.0 & 1 \\
& arm & 3.7 & {\tiny +}0.0 & 1 & 4.7 & {\tiny +}1.5 & 6 & 4.5 & {\tiny +}3.5 & 13 & 5.4 & {\tiny +}1.4 & 19 & 8.5 & {\tiny +}4.7 & 7 \\
& foot & 0.4 & {\tiny +}0.0 & 1 & 0.7 & {\tiny +}0.0 & 6 & 1.8 & {\tiny +}0.0 & 6 & 1.3 & {\tiny +}0.3 & 2 & 1.6 & {\tiny +}1.0 & 6 \\

\hline
\multicolumn{2}{c:}{mean (105 parts)}
& 9.0 & {\tiny +}0.6 & 6.9 & 12.4 &  {\tiny +}2.0 &  19.1 & 16.5 &  {\tiny +}2.6 &  24.0 & 17.8 &  {\tiny +}3.3 & 24.6 & 19.0 &  {\tiny +}5.2 &  18.8 \\
\multicolumn{2}{c:}{absolute GA mAP} & & 9.6 & & & 14.4 & & & 19.1 & & & 21.1 & & & 24.2 \\

\hline
\multicolumn{2}{c:}{all filters mAP} & 5.5 &  & 96 & 4.5 & & 256 & 5.6 & & 384 & 6.5 &  & 384 & 9.2 & & 256 \\
\hline
\end{tabular}}
\caption{\small Part detection results in terms of AP on the \texttt{\small train} set of PASCAL-Part for AlexNet-Object. \emph{Best} is the AP of the best individual filter whereas \emph{GA} indicates the increment over \emph{Best} obtained by selecting the combination of \emph{(nFilters)} filters.} 
\label{table:part_obj}}
\end{table*}

\subsubsection{Experimental settings}
\label{sec:cnn:exp_set}
\noindent{\it Dataset.}
We evaluate filters on the recent PASCAL-Part dataset~\citep{chen14cvpr}, which augments PASCAL VOC 2010~\citep{everingham10ijcv} with pixelwise semantic part annotations. For our experiments we fit a bounding-box to each part segmentation mask.
We use the \texttt{train} subset and evaluate all parts listed in PASCAL-Part with some minor refinements:
we discard fine-grained labels (e.g. `car wheel front-left' and `car wheel back-left' are both mapped to \emph{car-wheel}),
merge contiguous subparts of the same larger part (e.g. `person upper arm' and `person lower arm' become a single part \emph{person-arm}),
discard very tiny parts (average of widths and heights over the whole training set $\le 15$ pixels, like `person eyebrow'),
and discard parts with less than $\le 10$ samples in \texttt{train} (like 'bicycle headlight', which has only one annotated sample).
The final dataset contains 105 parts of 16 object classes. \\%

\noindent{\it AlexNet.}
One of the most popular networks in computer vision is the CNN model of Krizhevsky et al.~\citep{krizhevsky12nips}, winner of the ILSVRC 2012 image classification challenge~\citep{russakovsky15ijcv}. 
It is commonly referred to as AlexNet.
This network has 5 convolutional layers followed by 3 fully connected layers. The number of filters at each of the convolutional layers L is: 96 (L1), 256 (L2), 384 (L3), 384 (L4), and 256 (L5). The filter size changes across layers, from 11x11 for L1, to 5x5 to L2, and to 3x3 for L3, L4, L5.\\

\noindent{\it Training.}
We use the publicly available AlexNet network of \cite{girshick14cvpr}. The network was initially pre-trained for image classification on the ILSVRC12 dataset and subsequently finetuned for object class detection (for the 20 classes in PASCAL VOC 2012 $+$ background) using ground-truth annotations. Note how these bounding-boxes provide a coordinate frame common across all object instances. This makes it easier for the network to learn parts as it removes variability due to scale changes (the convolutional filters have fixed size) and presents different instances of the same part class at rather stable positions within the image. We refer to this network as {\it AlexNet-Object}. 
The network is trained on the \texttt{train} set of PASCAL VOC 2012. Note how this set is a superset of PASCAL VOC 2010 \texttt{train}, on which we analyze whether filters correspond to semantic parts. \\

\noindent Finally, we assist each of its filters by providing a bounding-box regression mechanism that refines its stimulus detections to each part class (sec.~\ref{sec:cnn:reg}) and we learn the optimal combination of filters for a part class using a GA (sec.~\ref{sec:cnn:GA}). \\

\noindent{\it Evaluating settings.}
We restrict the network inputs to ground-truth object bounding-boxes. More specifically, for each part class we look at the filter responses only inside the instances of its object class and ignore the background. For example, for \emph{cow-head} we only analyze \emph{cow} ground-truth bounding-boxes.
Furthermore, before inputting a bounding-box to the network we follow the R-CNN pre-processing procedure~\citep{girshick14cvpr}, which includes adding a small amount of background context and warping to a fixed size. An example of an input bounding-box is shown in fig.~\ref{fig:cnn:overview}. 
\rev{Finally, we intentionally evaluate on the \texttt{train} subset that was used to train the network (for object detection, without part annotations).}
These settings are designed to be favorable to the emergence of parts as we ignore image background that does not contain parts and use object instances seen by AlexNet-Object during training.

\subsubsection{Results}
\label{sec:cnn:resAP}

Table~\ref{table:part_obj} shows results for \rev{a} few parts of seven object classes in terms of average precision (AP). 
For each part class and network layer, the table reports
the AP of the best individual filter in the layer (`Best'), the increase in performance over the best filter thanks to selecting a combination of filters with our GA (`GA'), and the number of filters in that combination (`nFilters'). 
Moreover, the last three rows of the table report the mAP over all 105 part classes.
This includes the absolute performance of the GA combination in the second to last row, for easy reference, and the performance considering all filters simultaneously.
Several interesting facts arise from these results. \\

\noindent{\it Need for regression.}
In order to quantify how much the bounding-box regression mechanism of sec.~\ref{sec:cnn:reg} helps, we performed part detection using the non-regressed receptive fields. On AlexNet-Object layer 5, taking the single best filter for each part class achieves an mAP of 6.1. This is very low compared to mAP 19.0 achieved by assisting the filters with the regression.
Moreover, results show that the receptive field is only able to detect large parts (e.g. \emph{bird-torso, bottle-body, cow-torso}, etc.). This is not surprising, as the receptive field of layer 5 covers most of the object surface (fig.~\ref{fig:cnn:qualResRegression}).
Instead, filters with regressed receptive fields can detect much smaller parts (e.g. \emph{cat-ear, cow-muzzle, person-hair}),
as the regressor shrinks the area covered by the receptive field and adapts its aspect ratio to the one of the part.
Some filters at layer 1 benefit of the opposite effect. The reason why such filters with tiny receptive fields can detect large parts such as cow-torso and horse-torso, is because regression enlarges them appropriately. Without regression, layer 1 filters have zero AP for these parts.
We conclude that the receptive field alone cannot perform part detection and regression is necessary.

\noindent{\it Differences between layers.}
Generally, the higher the network layer, the higher the performance (table~\ref{table:part_obj}). 
This is consistent with previous observations~\citep{zeiler14eccv,zhou15iclr} that the first. layers of the network respond to generic corners and other edge/color junctions, while higher levels capture more complex structures. Nonetheless, it seems that some of the best individual filters of the very first layers can already perform detection to a weak degree when helped by our regression (e.g. \emph{bike-wheel} has 14.9 AP). \\

\noindent{\it Differences between part classes.}
Performance varies greatly across part classes. For example, some parts are clearly not represented by any filter nor filter combination, as their AP is very low across all layers
(e.g. 7.0 for \emph{aeroplane-engine}, 3.6 for \emph{bike-chainwheel}, 2.6 AP for \emph{person-foot}, at layer 5 using the GA combinations).
On other parts instead, the network achieves good detection performance
(e.g. 64.2 for \emph{bike-wheel}, 59.1 for \emph{cat-head}, and 72.3 AP for \emph{horse-torso}).
This proves that some of the filters can be associated with these parts. \\

\begin{figure}
  \begin{center}
    \includegraphics[width=\textwidth]{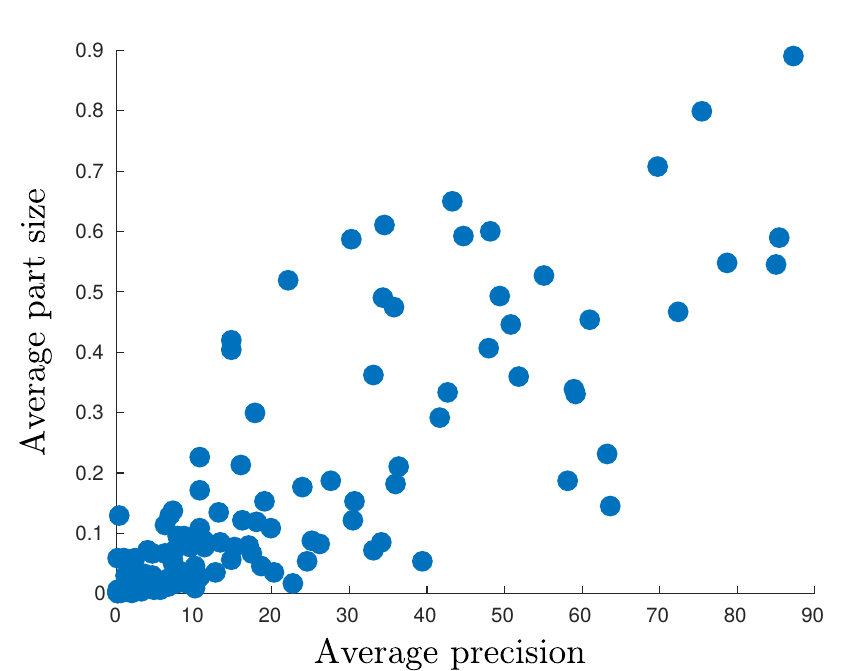}
\end{center}
\caption{\small Correlation between the size of a part class, averaged over all its instances,
  and part detection performance (AP for the best combination of layer 5 filters found by GA, table~\ref{table:part_obj}).
The part size is normalized by the average size of the object class it belongs to.
Each point corresponds to a different part class. These are highly correlated: Pearson product-moment correlation coefficient of 0.7.} 
\label{fig:cnn:APvsPartSize}
\end{figure}

\noindent{\it Differences between part sizes.}
Another factor that seems to influence the performance is the average size of the part.
For example, the AP achieved on \emph{horse-torso} (72.3 on layer 5) is much higher than on the smaller \emph{horse-ear} (6.6).
In order to understand if this is common across all parts, we looked at how AP changes with respect to the average size of a part (fig.~\ref{fig:cnn:APvsPartSize}). 
Interestingly, these two are indeed correlated and have a Pearson product-moment correlation coefficient (PPMCC) of 0.7~\citep{pearson1895jstor}. 
This shows that smaller parts emerge less in the CNN than larger ones.
Nonetheless, small size does not always imply low detection performance: there are some rather small parts (around 20\% of the object area) which have high AP (around 60), like \emph{bicycle-wheel}, \emph{motorbike-wheel} and \emph{person-head}.
As we show in sec.~\ref{sec:cnn:discrim_parts}, these are parts that are very discriminative for recognizing the objects they belong to, which justifies their emergence within the network.\\

\begin{figure*}
  \begin{center}
    \includegraphics[width=0.9\textwidth]{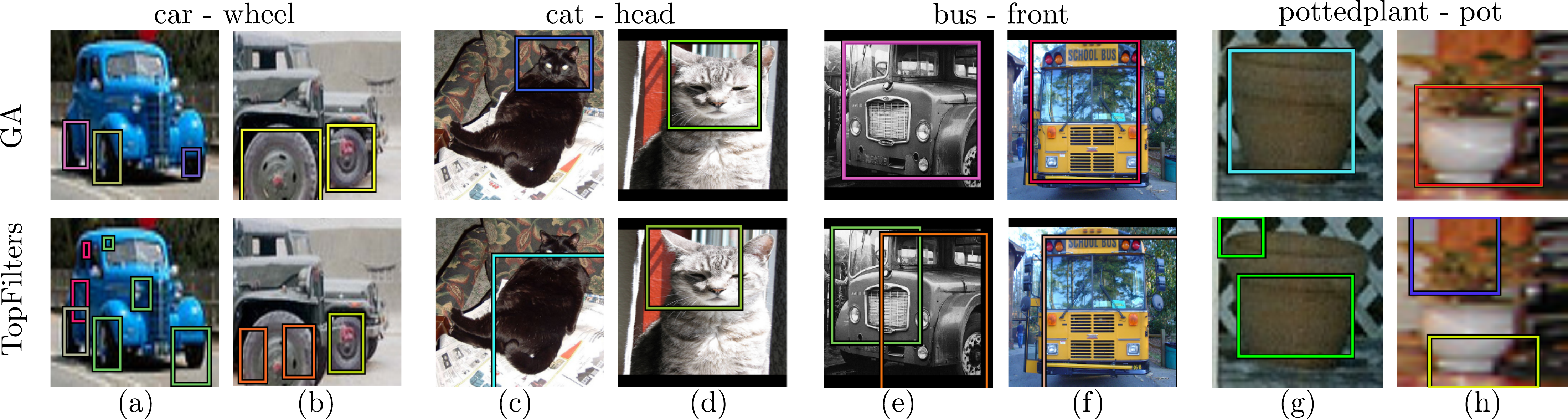}
\end{center}
\caption{\small Part detection examples obtained by combination of filters selected by our GA (top) or by TopFilters (bottom). Different box colors correspond to different filters' detections. Note how the GA is able to better select filters that complement each other. } 
\label{fig:cnn:GAvsTop_vis}
\end{figure*}

\begin{figure}
  \begin{center}
    \includegraphics[width=.9\textwidth]{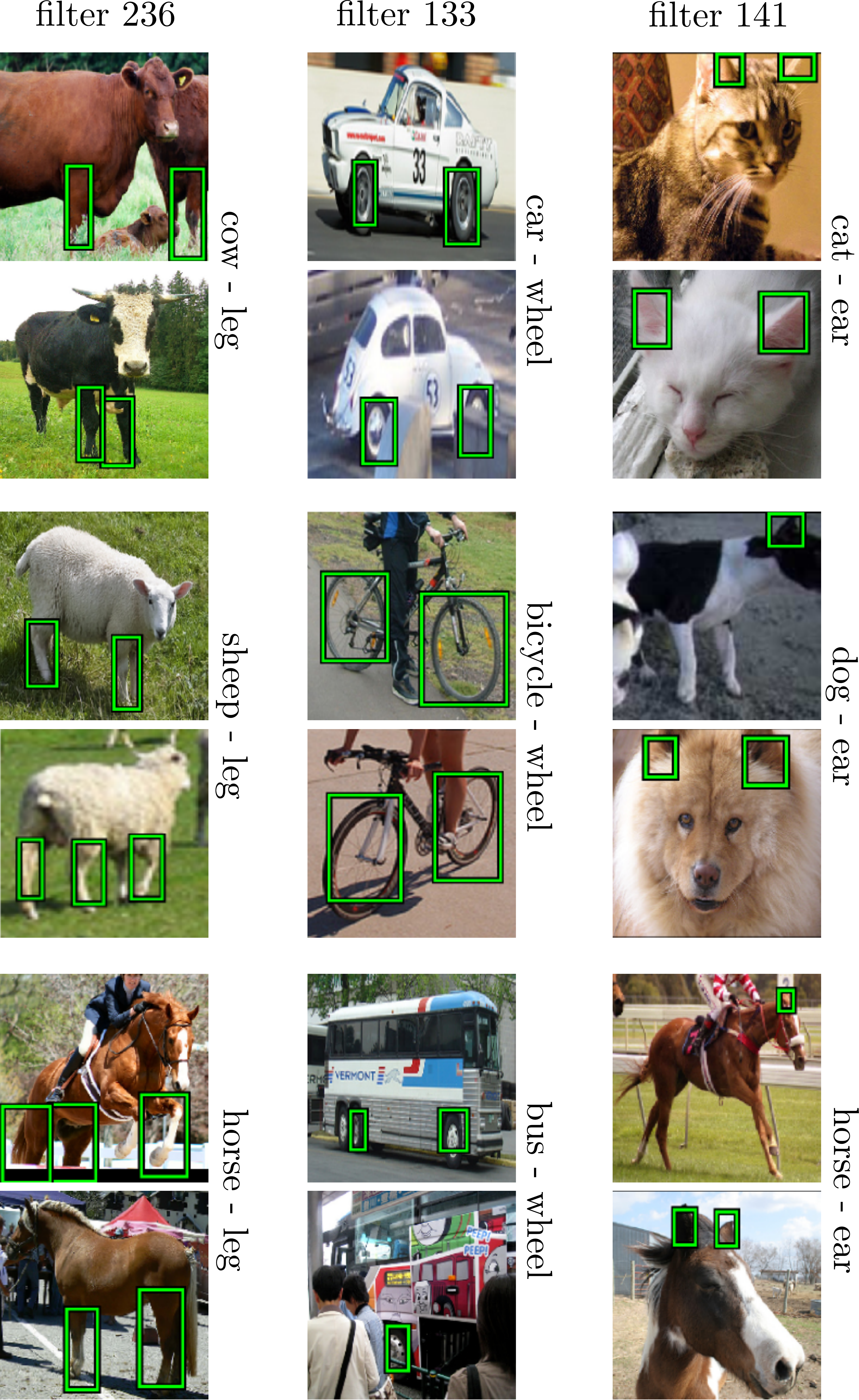}
\end{center}
\caption{\small Detections performed by filters 141, 133, and 236 of AlexNet-Object, layer 5. The filters are specific to a part and they work well on several object classes containing it. }
\label{fig:cnn:shared_filters_parts}
\end{figure}

\noindent{\it Filter combinations by GA.}
Performing part detection using a combination of filters (GA) always performs better (or equal) than the single best filter. This is interesting, as it shows that different filters learn different appearance variations of the same part class. 
On the other hand, simultaneously using all filters brings a very low performance. 
This is due to irrelevant filters producing many false positives, which reduces AP.
 Moreover, combining multiple filters selected by the GA improves part detection performance more for deeper layers. 
 This suggests that they are more class-specific, i.e. they dedicate more filters to learning the appearance of specific object/part classes. This can be observed by looking not only at the improvement in performance brought by the GA, but also at the number of filters that the GA selects. Clearly, filters in L1 are so far from being parts that even selecting many filters does not bring much improvement (+0.6 mAP only). 
Instead, in L4 and L5 there are more semantic filters and the GA combination helps more (+3.3 mAP and +5.2 mAP, respectively). Interestingly, for L5 the improvement is higher than for L4, yet fewer filters are combined. 
This further shows that filters in higher layers better represent semantic parts.

\noindent{\it TopFilters: a baseline for combining filters.}
The AP improvement provided by our GA for some parts is remarkable, like for \emph{aeroplane-body} (+17.0), \emph{horse-leg} (+9.6) and \emph{cow-head} (+17.1).

These results suggest that our GA is doing a good job in selecting filter combinations.
Here we compare against a simpler method, dubbed TopFilters.
It selects the top few best filters for a part class, based on their individual AP.
We let TopFilters select the same number of filters as the GA.
Our GA consistently outperforms TopFilters (24.2 vs 18.8 mAP,
layer 5). 
The problem with TopFilters \rev{seems to be} that often the top individual best filters capture the same visual aspect of a part. Instead, our GA can select filters that complement each other and work well jointly (indeed 57\% of the filters it selects are not among those selected by TopFilters). 
We can see this phenomenon in fig.~\ref{fig:cnn:GAvsTop_vis}.
On the blue car, TopFilters detects two wheels correctly, but fails to fit a tight bounding-box around the third wheel that appears much smaller (fig.~\ref{fig:cnn:GAvsTop_vis}{\color{blue}a}).
Similarly, on the other car TopFilters fails to correctly localize the large wheel (fig.~\ref{fig:cnn:GAvsTop_vis}{\color{blue}b}). Instead, the GA localizes all wheels correctly in both cases.
Furthermore, the GA fits tighter bounding-boxes for more challenging parts, achieving more accurate detections (fig.~\ref{fig:cnn:GAvsTop_vis}{\color{blue}c-h}).
Finally, note how TopFilters does not even improve over selecting the single best filter (19.0), as more filters bring more false positive detections.

\noindent{\it GA convergence.}
We present here an experiment to study the convergence of the GA by assessing the variance of its solutions.
We take \emph{car-wheel} as example part class since it has reasonable accuracy.
We ran the GA at layer 5 for 10 trials starting with different random seeds.
On average, each trial selects 16 filters, with a standard deviation of 2.7.
The AP remains very similar for all runs (mean 39.42, standard deviation of 0.03).
This indicates that, for the task at hand, our GA is stable, converging to equivalent solutions in terms both of the number of selected filters and their collective performance.

\noindent{\it Filter sharing across part classes.}
We looked into which filters were selected by our GA and noticed that some are shared across different part classes.
We then confirmed that those filters have high part detection performance for equivalent part classes across different object classes (e.g. \emph{car-wheel} and \emph{bicycle-wheel}). 
Fig.~\ref{fig:cnn:shared_filters_parts} shows some examples of these filters' detections.
It is clear that some filters are representative for a generic part and work well on all object classes containing it.

\begin{figure}
  \begin{center}
      \includegraphics[width=0.95\textwidth]{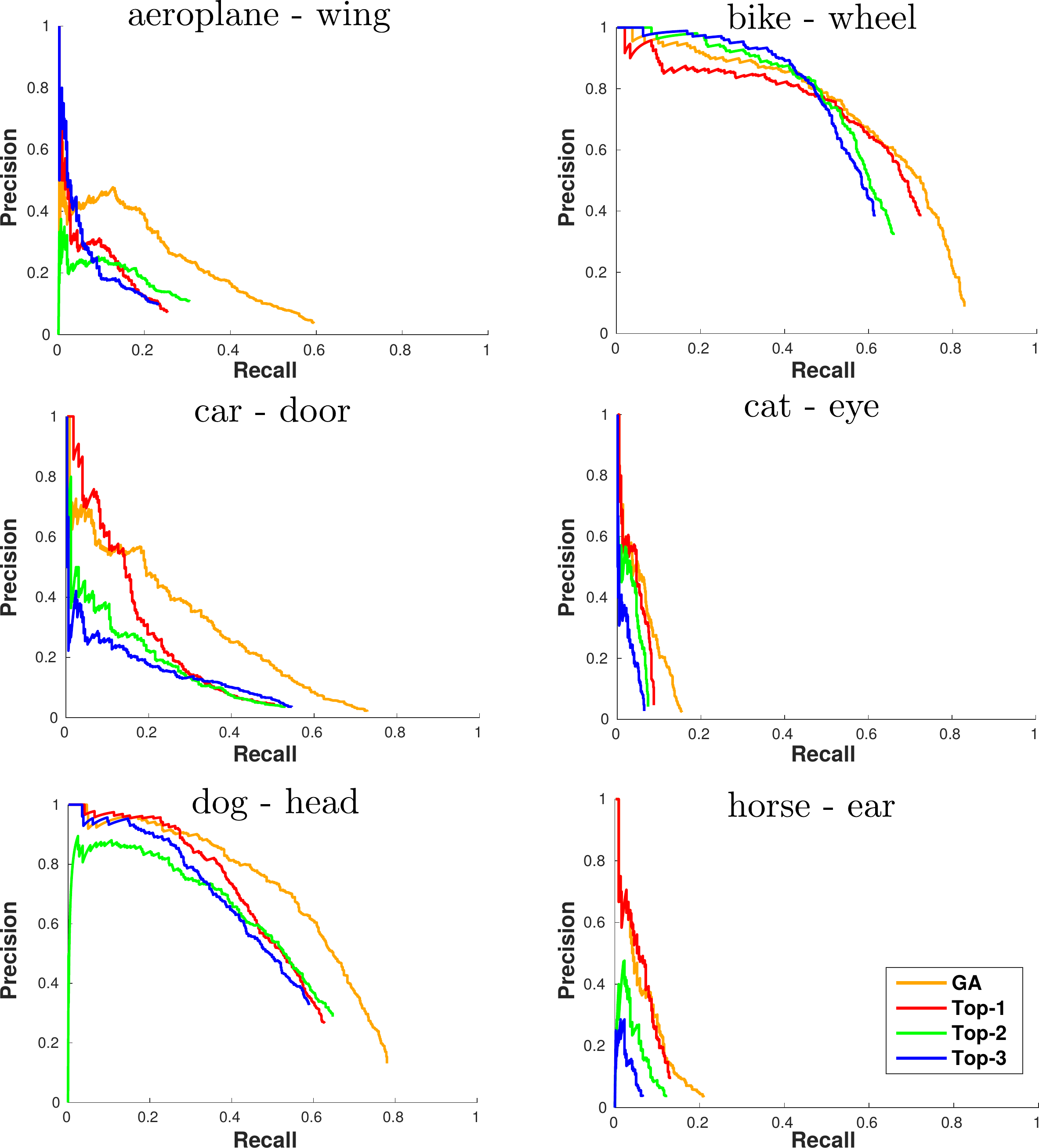}
  \end{center}
  \caption{\small Precision vs. recall curves for six part classes using AlexNet-Object's layer 5 filters. For each part class we show the curve for the the top three individually best filters and for the combination of filters selected by our GA. }
  \label{fig:cnn:coverage}
\end{figure}

\noindent{\it Instance coverage.}
Table~\ref{table:part_obj} shows high AP results for several part classes, showing how some filters can indeed act as part detectors.
However, as AP conflates both recall and precision, it does not reveal how many part instances the filters cover.
To answer this question, fig.~\ref{fig:cnn:coverage} shows precision vs. recall curves for several part classes.
For each part class, we take the top three filters of layer 5, and compare them to the filter combination returned by the GA.
We can see how the combination reaches higher AP not only by having higher precision (fewer false positives) in the low recall regime, but also by reaching considerably higher recall levels than the individual filters.
For some part classes, the filter combination covers as many as 80\% of its instances (e.g. \emph{car-door, bike-wheel, dog-head}).
For the more challenging part classes, neither the individual filters nor the combination achieve high recall levels, suggesting that the convolutional filters have not learned to respond to these parts systematically (e.g. \emph{cat-eye, horse-ear}). \\

\noindent{\it How many semantic parts emerge in AlexNet-Object?}
So far we discussed part detection performance for individual filters of AlexNet-Object and their combinations.
Here we want to answer the main bottomline question: for how many part classes does a detector emerge? We answer this for two criteria: AP and instance coverage.

For AP, we consider a part to emerge if the detection performance for the best filter combination in the best layer (L5) exceeds 30 percent AP.
This is a rather generous threshold, which represents the level above which the part can be somewhat reliably detected.
According to this criterion, 34 out of the 105 semantic part classes emerge.
This is a modest number, despite all favorable conditions we have engineered into the evaluation and all assists we have given to the network (including bounding-box regression and optimal filter combinations).

According to the instance coverage criterion, instead, results are more positive. We consider that a filter combination covers a part when it reaches a recall level above 50\%, regardless of false-positives. According to this criterion, 71 out of the 105 part classes are covered, which is greater than the number of part detectors found according to AP.
This indicates that, although for many part classes there is a filter combination covering many of its instances, it also fires frequently on other image regions (leading to high false \rev{positive} rates).

Based on all this evidence, we conclude that the network does contain filter combinations that can cover some part classes well, but they do not fire exclusively on the part, making them weak part detectors. 
This demystifies the observations drawn through casual visual inspection, as in~\citep{zeiler14eccv}.
Moreover, the part classes covered by such semantic filters tend to either cover a large image area, such as torso or head, or be very discriminative for their object class, such as wheels for vehicles and wings for birds.
Most small or less discriminative parts are not represented well in the network filters, such as headlight, eye or tail.

\subsection{Other network architectures and levels of supervision}
\label{sec:cnn:alexnet_others}

We now explore how the level of supervision provided during training and the network architecture affect what the filters learn.\\

\noindent{\it Networks and training.}
We consider several additional networks with different supervision levels (\emph{AlexNet-Image}, \emph{AlexNet-Scenes}, and \emph{AlexNet-Object$\leftarrow$Scratch}), and a different architecture (\emph{VGG16-Object}). 

AlexNet-Image~\citep{krizhevsky12nips} is trained for image classification on 1.3M images of 1000 object classes in ILSVRC 2012~\citep{russakovsky15ijcv}. 
Note how this network has not seen object bounding-boxes during training. 
For this reason, we expect its filters to learn less about semantic parts than AlexNet-Object.
On the opposite end of the spectrum, AlexNet-Scene~\citep{zhou14nips} is trained for scene recognition on 205 categories of the Places database~\citep{zhou14nips}, which contains 2.5M scene-centric images. 
As with AlexNet-Image, \rev{AlexNet-Scene} has not seen object bounding-boxes during training. But now the training images show complex scenes composed of many objects, instead of focusing on individual objects as in ILSVRC 2012.
Moreover, while the network might learn to use objects as cues for scene recognition~\citep{zhou15iclr}, the task also profits from background patches (e.g. \emph{water} and \emph{sky} for beach). For these reasons, we expect object parts to emerge even less in AlexNet-Scene.
For both AlexNet-Image and AlexNet-Scene we use the publicly available models from \citep{jia13caffe}. 

We introduce AlexNet-Object$\leftarrow$Scratch in order to assess the importance of pre-training in AlexNet-Object. 
We directly train this network for object detection on PASCAL VOC 2012 from scratch, i.e. randomly initializing its weights instead of pre-training on ILSVRC 2012. The rest of the training process remains identical to the one for AlexNet-Object (sec.~\ref{sec:cnn:exp_set}).

Finally, VGG16-Object is the 16-layer network of \cite{simonyan15iclr}, finetuned for object detection~\citep{girshick14cvpr} on PASCAL VOC 2012 (like AlexNet-Object).
While its general structure is similar to AlexNet, it is deeper and the filters are smaller (3x3 in all layers), leading to better image classification~\citep{simonyan15iclr} and object detection~\citep{girshick15iccv} performance. 
Its convolutional layers can be grouped in 5 blocks.
The first two blocks contain 2 layers each, with 64 and 128 filters, respectively.
The next block contains 3 layers of 256 filters.
Finally, the last 2 blocks contain 3 layers of 512 filters each. \\

\begin{table}[t]
\centering
\footnotesize
\resizebox{\columnwidth}{!}{
\begin{tabular}{ C{2.4cm} : C{1.7cm} L{1.4cm} : C{0.7cm} C{0.7cm} C{0.7cm}} 
\toprule
\multirow{3}{*}{\bf Network name} & \multicolumn{2}{c:}{\multirow{2}{*}{\bf Training}} & \multicolumn{3}{c}{\multirow{2}{*}{\bf Results - Layer}} \\
& & & & & \\
& Pre-train & Train & \bf{L3} & \bf{L4} & \bf{L5} \\ 
\midrule
AlexNet-Object & \multicolumn{1}{c}{ILSVRC12}&VOC12 & 19.1 & 21.1 & 24.2\\
AlexNet-Image  & -&ILSVRC12 & 20.0 & 21.5 & 23.2\\
AlexNet-Scene & -&Places & 17.5 & 17.6 & 18.1\\
\midrule
AlexNet-Object$\leftarrow$S & \multicolumn{1}{c}{-}&VOC12 & 14.5 & 16.2 & 18.2\\
\midrule
VGG16-Object & \multicolumn{1}{c}{ILSVRC12}&VOC12 & 12.9 & 21.5 & 26.1 \\
VGG16-Image & \multicolumn{1}{c}{-} & \multicolumn{1}{c:}{ILSVRC12} & 12.4 & 19.6 & 22.1 \\
\bottomrule
\end{tabular}}
\caption{\small Part detection results (mAP). For VGG-16, L3, L4, and L5 correspond to L3\textunderscore 3, L4\textunderscore 3, and L5\textunderscore 3, respectively. }
 \label{table:conv_sup}
\end{table}

\noindent{\it Results.}
Table~\ref{table:conv_sup} presents results for all networks we consider.
For the AlexNet architectures, we focus on the last three convolutional layers, as we observed in sec.~\ref{sec:cnn:resAP} that filters in the first two layers correspond poorly to semantic parts. 
Analogously, for VGG16-Object we present the top layer of each of the last 3 blocks of the network (L3\textunderscore3, L4\textunderscore3, and L5\textunderscore3). 
We report mAP results obtained by the GA filter combination, averaged over all part classes.

Both AlexNet-Image and AlexNet-Object present reasonable part emergence across all their layers.
This shows that the network's inclination to learn semantic parts is somewhat already present even when trained for whole image classification, suggesting that object parts are useful for that task too.
However, the part emergence on L5 for AlexNet-Object is higher.
This indicates that parts become more important when the network is trained for object detection, affecting particularly higher layers, near the final classification layer that can use the responses of these filters to recognize the object.

Interestingly, parts emerge much less when training the network for scene recognition, as the results of AlexNet-Scene indicate (-6.1\% mAP compared to AlexNet-Object).
The relative performance of the three networks AlexNet-Object, AlexNet-Image, AlexNet-Scene suggest that the network seems to learn parts to the degree it needs them for the task it is trained for, a remarkable behaviour indeed.

AlexNet-Object$\leftarrow$Scratch performs clearly worse than AlexNet-Object, which is likely due to the fact that PASCAL VOC 2012 is too small for training a complex CNN, and so pre-training on ILSVRC is necessary~\citep{agrawal14eccv,girshick14cvpr}.
Finally, parts emerge more in the deeper VGG16-Object than in AlexNet-Object (L5). 
  The higher part emergence could be due to the better performance of this network or its greater depth.
  In order to investigate this, we also test VGG16-Image: VGG16 architecture just pre-trained for image classification, like AlexNet-Image.
  This model preserves the depth but loses the performance advantage over the fine-tuned counterparts.
  The results are similar to AlexNet-Image for layers 4 and 5, suggesting that the better performance is indeed responsible for the higher part emergence.
  However, there is still correlation with depth as better networks tend to be deeper.

All the networks but AlexNet-Scene confirm the trend observed for AlexNet-Object: filters in higher layers are more responsive to semantic parts.
This is especially notable for VGG16-Object.
As this network has many more layers, the levels of semantic abstraction are more spread out.
For example, L3\textunderscore3 has very low emergence as there are six more layers above it instead of two in AlexNet.
Therefore, the network can postpone the development of semantic part filters to later layers.
AlexNet-Scene displays the same, rather low level of responsiveness to semantic parts in all layers considered. We hypothesize this is due to semantic object parts playing a smaller role for scene recognition.

\section{Semantic part emergence in CNNs according to humans}
\label{sec:cnn:humans}
Our quantitative evaluation presented in sec.~\ref{sec:cnn:pascalParts} uses the semantic part annotations available in PASCAL-Part dataset \citep{chen14cvpr} to determine how many semantic parts emerge in the CNNs.
We address now the converse question: \emph{``what fraction of all filters respond to any semantic part?''}
Despite being the best existing parts dataset, PASCAL-Part is not complete: some semantic parts are not annotated (e.g. the door handle of a car).
For this reason, we cannot answer this new question using it, as a filter might be responding to an unannotated semantic part.

We propose here a human-based experiment that goes beyond the semantic parts annotated in PASCAL-Part.
For each object class we ask human annotators if activations of a filter systematically correspond to a semantic part of that object class, and, if yes, to name the part (sec.~\ref{sec:cnn:humans:meth}).
This data provides a mapping from filters to \rev{semantic} parts, which is only limited by the semantic parts known by the annotator.
Using this mapping, we can now answer the proposed question (sec.~\ref{sec:cnn:humans:res}).
Moreover, this mapping also allows us to compare the parts emerging in this human experiment with the parts that emerged according to the PASCAL-Part annotations(sec.~\ref{sec:cnn:pascalParts}).
This enables to discern whether some other parts emerge besides those annotated in PASCAL-Part (sec.~\ref{sec:cnn:humans:comp_pascal}).

\begin{figure}[t]
  \begin{center}
    \includegraphics[width=\textwidth]{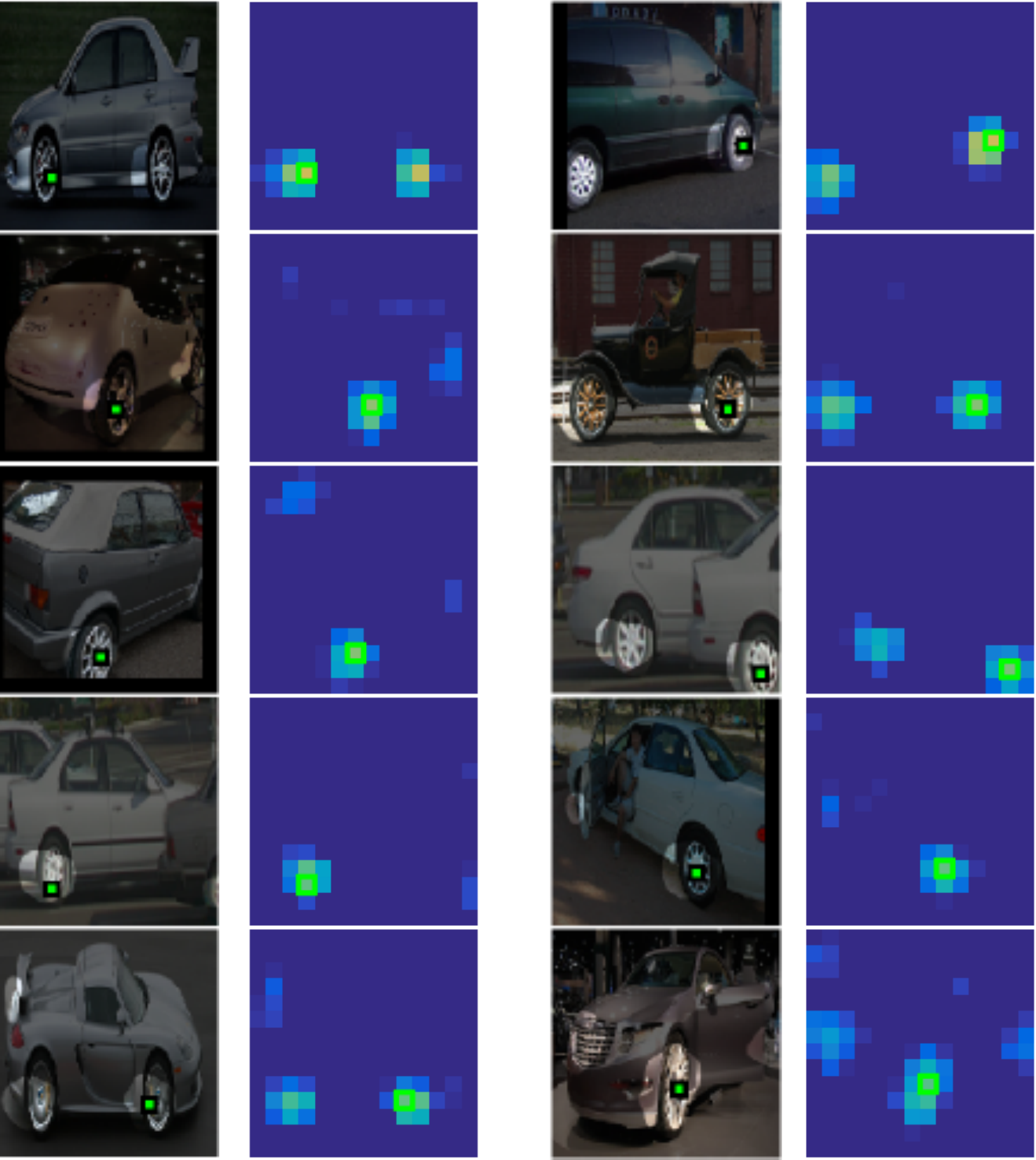}
  \end{center}
  \caption{\small Example of a question shown to our annotators, filter id: 186, class: \emph{car}. We show the top 10 images with the highest activation of the filter on this object class, along with the corresponding activation maps.}
  \label{fig:cnn:exampleFilterAct}
\end{figure}

\subsection{Methodology}
\label{sec:cnn:humans:meth}
For each pair of object class and filter, we present an image like fig.~\ref{fig:cnn:exampleFilterAct} to an annotator.
The image shows the top 10 activations of the filter on object instances of the class (\emph{car}, in the figure).
We show the image shaded and overlay the activation map by setting the transparency value of the shading proportionally to the activation value at a pixel. This highlights the activation map and helps the annotator to quickly see high activation regions in the context of the rest of the image.
We also indicate the maximum of the activation map with a green square to emphasize the maximum activation (which is typically in the middle of the region). 

The task consists in answering the question: \emph{``Do the highlighted areas in the images systematically cover  the same concept, and if so, which?''}. 
This is the case if the highlighted areas cover the same concept in at least seven out of the ten images.
The possible answers are the following:
\begin{enumerate}
  \item Yes - Semantic part 
  \item Yes - Background
  \item Yes - Other 
  \item No 
  \item Not sure
\end{enumerate}

\begin{figure*}
  \begin{center}
    \includegraphics[width=0.95\textwidth]{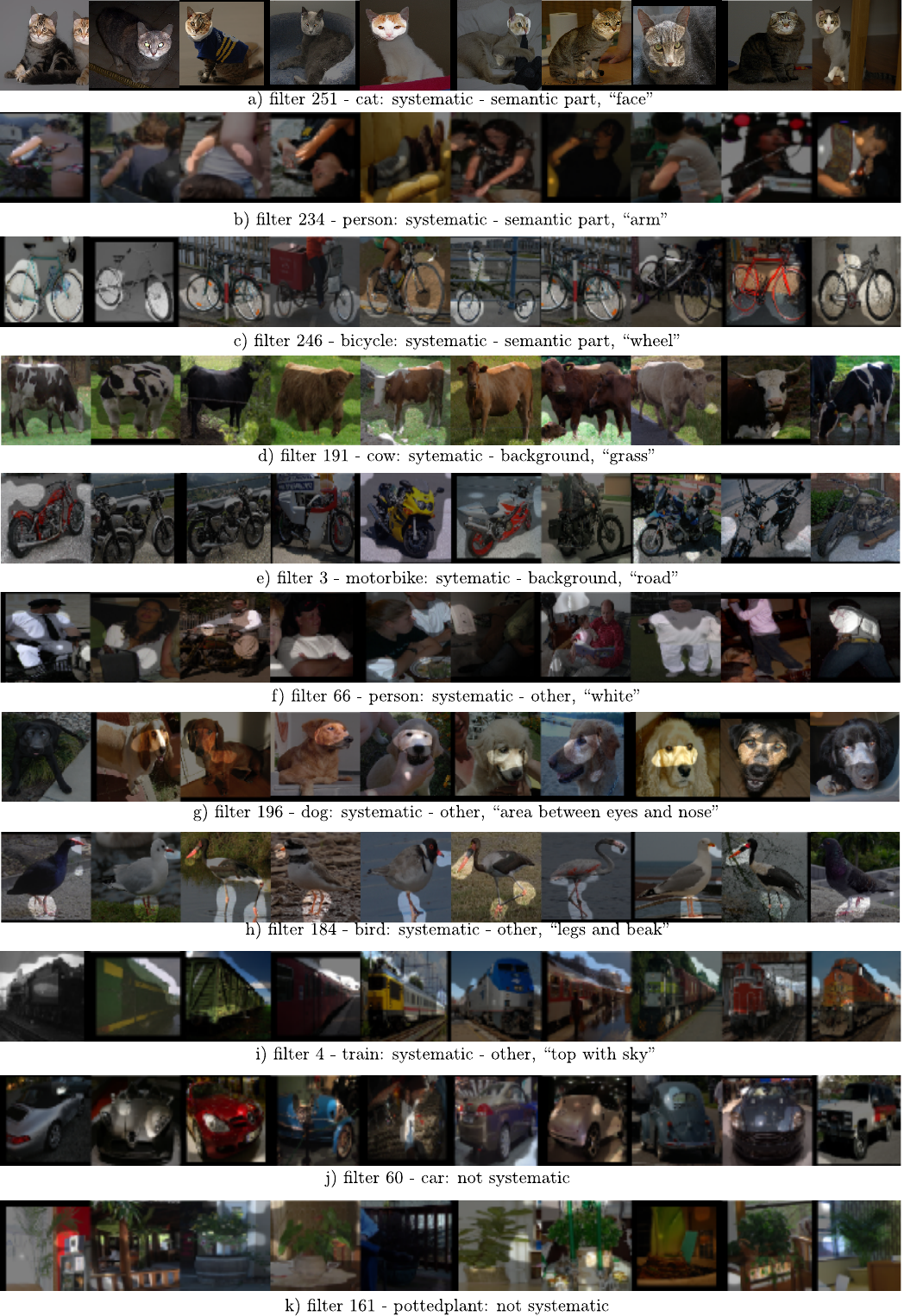}
\end{center}
\caption{\small \rev{Examples} of human annotation for different filters and classes.}
\label{fig:cnn:humanExamples}
\end{figure*}

In case of an affirmative response, the annotator needs to specify one of three types of concepts: semantic part (e.g. wheel), background (e.g. grass) or anything else (e.g. white color). Additionally, we ask them to name the concept by typing it in a free-text field.
The idea behind this protocol is to distinguish filters that fire on a variety of different image structures (fig.~\ref{fig:cnn:humanExamples}{\color{blue}j-k}), including occasionally some semantic parts, from genuine part detectors, which fire systematically on a particular part.  
Furthermore, we have expanded our experiment beyond semantic parts (i.e. background and other) in order to achieve a more comprehensive understanding of the filter stimuli.
The last option (``Not sure'') allows the annotator to skip ambiguous cases, which we later reject. 

We ask a question for each combination of object class and network filter.
We explore L5 filters of AlexNet-Object (256) and we consider the 16 object classes used in sec.~\ref{sec:cnn:pascalParts}, leading to a total of $256 \times 16$ questions. 
We use two expert annotators to process half of the object classes each.
To measure agreement, they also process one of the object classes from the other annotator's set.
Their agreement on the types of filters is high: in 79\% of the questions, the two annotators clicked on the same answer (out of the 5 possible answers above).

\begin{figure}[t]
  \begin{center}
    \includegraphics[width=\textwidth]{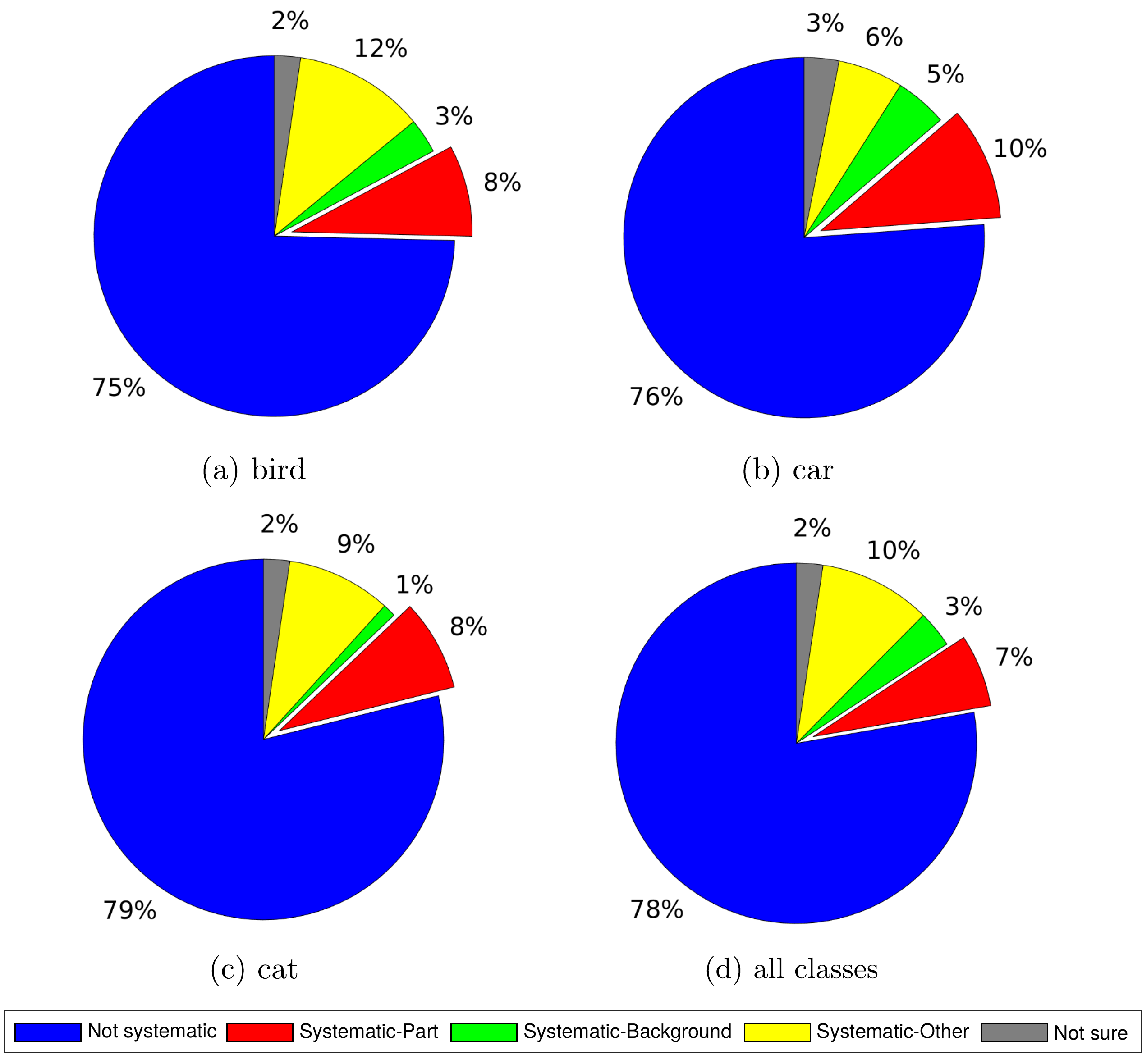}
  \end{center}
  \caption{\small Filter distributions for bird, car, cat, and the average of all 16 classes.} 
  \label{fig:cnn:filterDist}
\end{figure}

\subsection{Results}
\label{sec:cnn:humans:res}

This experiment enables to \rev{obtain} a distribution over the types of filters for each object class.
Fig.~\ref{fig:cnn:filterDist} shows these distributions for three example object classes (\emph{bird}, \emph{car}, and \emph{cat}) as well as the average result for all 16 object classes. 
Additionally, fig.~\ref{fig:cnn:humanExamples} shows some example human answers.
The majority of the filters do not seem to systematically respond to any identifiable concept (fig.~\ref{fig:cnn:humanExamples}{\color{blue}j-k}).
Among the filters that do respond to concepts systematically, only an average of 7\% (18 filters) correspond to semantic parts of a particular object class (fig.~\ref{fig:cnn:humanExamples}{\color{blue}a-c}).
Only 3\% of the filters respond systematically to background patches, examples include ``grass'', ``road'',  and ``sky''(fig.~\ref{fig:cnn:humanExamples}{\color{blue}d-e}).
Finally, most of the systematic filters, 10\% overall, respond to some other concepts. 
Among these, we most often find colors and textures (fig.~\ref{fig:cnn:humanExamples}{\color{blue}f}), but also subregions of a part (fig.~\ref{fig:cnn:humanExamples}{\color{blue}g}), assemblies of multiple semantic parts or their subregions (fig.~\ref{fig:cnn:humanExamples}{\color{blue}h}), or even regions straddling between a part and a background patch (fig.~\ref{fig:cnn:humanExamples}{\color{blue}i}). 

By further inspection, we found that background filters are consistent across images of different object classes.
For example, a filter that systematically fires on ``grass'' patches does it for most classes commonly found outdoors.
Similarly, some of the ``other'' systematic filters, especially the ones responding to colors, also exhibit the same behavior across object classes.
In contrast, the situation for systematic filters responding to semantic parts is mixed.
Although in some cases a filter responds to similar semantic parts of different object classes (like ``wheel'' or ``leg'', as in fig.~\ref{fig:cnn:shared_filters_parts}), in some other cases this does not hold.
For example, there is a filter that responds to ``wheel'' (in \emph{car} images), ``leg'' (in \emph{cow} images), and ``paw'' (in \emph{cat} images).
This indicates that the filter is responding to several types of stimuli simultaneously, possibly due to a higher order stimulus of which humans are not aware.

\begin{table}
\centering
\tiny
\resizebox{.7\columnwidth}{!}{
\begin{tabular}{l l} 
  \toprule
aeroplane & nose \\ 
bicycle & frame, hub, spokes, tire, tube \\ 
bird & belly \\ 
bottle & base, finish, neck, shoulder \\ 
bus & hood \\
car & fender, grill \\
cat & back \\
cow & belly \\
dog & forehead  \\
horse & belly, forehead, shoulder \\
motorbike & rim \\
person & crotch \\
pottedplant & pot rim, soil \\
sheep &  belly \\
train & engine, headlight, window \\
tv monitor & -\\
\bottomrule
\hline
\end{tabular}}
\caption{\small List of semantic parts that emerge in AlexNet-Object (layer 5) according to our human experiment, but that are not annotated in the PASCAL-Part dataset.}
\label{table:partsNotInPASCALP}
\end{table}

\subsection{Comparison to PASCAL-Part}
\label{sec:cnn:humans:comp_pascal}

In this section we compare the part emergence observed in sec.~\ref{sec:cnn:pascalParts} with the part emergence from this human experiment. Moreover, we also look at what semantic parts emerge according to our annotators, but are not present in \rev{PASCAL-Part}.\\

\paragraph{Emergence of PASCAL-Part classes.}
In sec.~\ref{sec:cnn:pascalParts} we observed that 34 out of the 105 semantic parts of PASCAL-Part emerge in AlexNet-Object according to our AP criterion (layer 5, sec. ~\ref{sec:cnn:resAP}).
24 out of these 34 parts also emerge according to the human judgements. Of the missing 10, four are animals \emph{torso}, for which humans prefer more localized names like ``back'' and ``belly'', four are vehicles viewpoints rather than actual semantic parts (e.g. \emph{bus-leftside} and \emph{car-frontside}). The remaining two are \emph{aeroplane-stern} and \emph{bottle-body}.
Hence, nearly all of the actual semantic parts that emerged according to detection AP also emerge according to human judgements. 

Overall, 59 of the semantic parts annotated in \rev{PASCAL-Part} emerge according to human judgements.
This is substantially more than the 34 that emerged according to detection AP.
The reason lies on the fact that it is easier for a filter to count as a semantic part in the human experiment, because it is tested only on the 10 images with the highest activations per object class (fig. \ref{fig:cnn:exampleFilterAct}). 
The AP criterion is more demanding:
it takes into account all instances of the part in the dataset,
it also counts false-positive detections, and a detection has to be spatially quite accurate to be considered correct (IoU$\geq0.4$). This might be a reason why works based on looking at responses on a few images, such as \citep{zeiler14eccv}, claimed that filters correspond to semantic parts: they only observed a few strong activations in which this happens.
\rev{
Our analysis goes a step further, by examining how the filters behave over the entire dataset.}

\paragraph{Emergence of other semantic part classes.}
In our human experiment, annotators are free to recognize and name \emph{any} semantic part. 
Table~\ref{table:partsNotInPASCALP} lists the 29 semantic parts that emerge in AlexNet-Object according to our annotators, but that are not annotated in PASCAL-Part.
Interestingly, 9 of them concentrate on two object classes: \emph{bicycle} (5) and \emph{bottle} (4). This can be explained by two observations. First, the new parts of the bicycle are mostly sub-parts on the wheel, which is the most discriminative part for the detection of the object (sec.~\ref{sec:cnn:discr_filters}). And second, the new parts of the bottle are all sub-part of the \emph{bottle-body} part as annotated in PASCAL-Part. As \emph{bottle-body} is essentially the whole object, the network prefers to learn finer-grained, actual parts.

Furthermore, two semantic parts often emerging in animal classes are ``belly'' and ``back''. Their emergence shows again how the network prefers more localized parts, rather than a larger ``torso'', as in the PASCAL-Part annotations.
Finally, we hypothesize that many of the remaining parts emerge because of their distinctive shapes. For example, \emph{aeroplane-nose} resemble a cone, and \emph{person-crotch} a triangle, \emph{car-fender} a semi-circle and \emph{motorbike-rim} a circle. Moreover, \emph{car-grill} and \emph{train-window} have a characteristic grid pattern.

\section{Discriminativeness of filters for object recognition}
\label{sec:cnn:discr_filters}

\begin{figure}[t]
  \begin{center}
    \includegraphics[width=\textwidth]{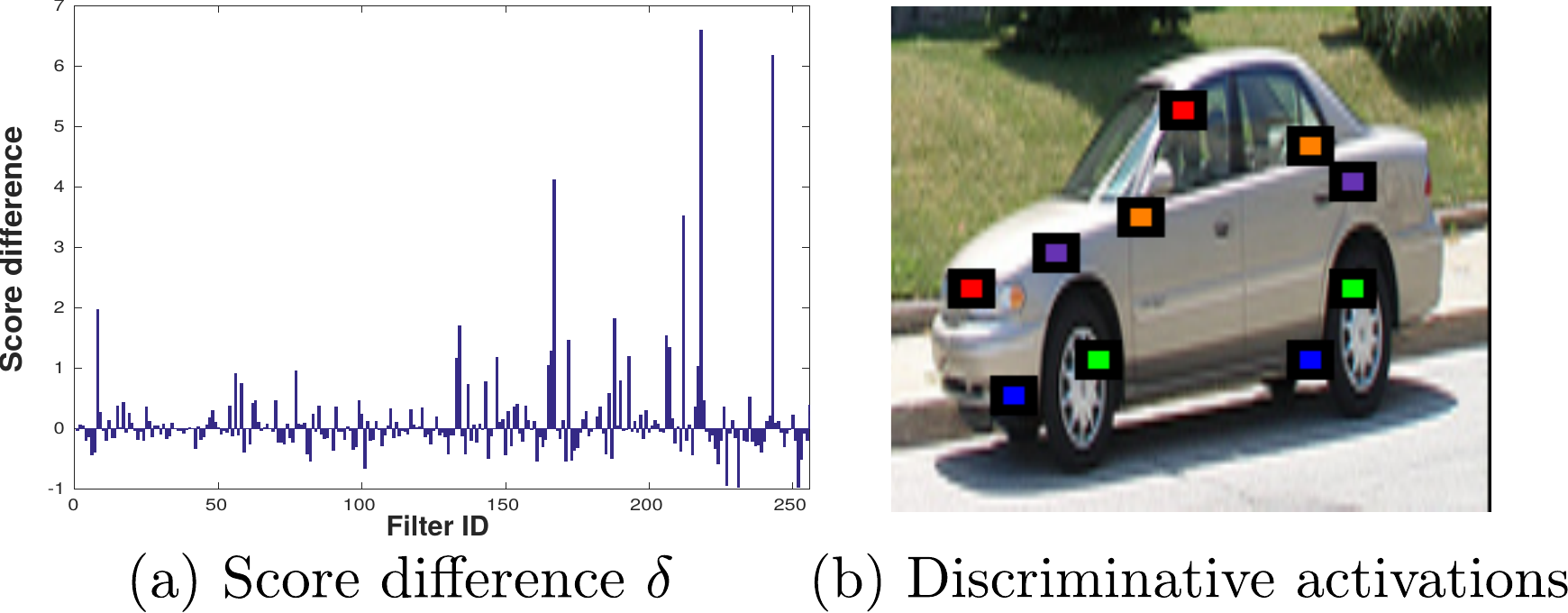}
  \end{center}
  \caption{\small Discriminative filters for object class {\em car}.
  (a) Shows how discriminative the filters of AlexNet-Object (layer 5) are for car (higher values are more discriminative).
  (b) Shows the activations of the five most discriminative filters on an example image. }
  \label{fig:cnn:scoreDiff}
\end{figure}

The training procedure of the CNNs we considered maximizes an objective function related to recognition performance, e.g. image classification or object detection.
Therefore, the network filters are likely to learn to respond to image patches discriminative for the object classes in the training set.
However, these discriminative filters need not correspond to semantic parts.
In this section we investigate to which degree the network learns such discriminative filters.
We investigate whether layer 5 filters of AlexNet-Object respond to recurrent discriminative image patches, by assessing how discriminative each filter is for each object class.
We use the following measure of the discriminativeness of a filter $f_j$ for a particular object class.
First, we record the output score $s_i$ of the network on an input image $I_i$.
Then, we compute a second score $s_i^j$ using the same network but ignoring filter $f_j$. 
We achieve this by zeroing the filter's feature map $x_j$, which means $a_{c,r} = 0$, $\forall a_{c,r} \in x_j$.
Finally, we define the discriminativeness of filter $f_j$ as the score difference averaged over the set $I$ of all images of the object class
\begin{equation}
  \small
  \delta_j = \dfrac{1}{|I|}\sum_{I_i\in I} s_i-s_i^j.
  \label{eq:discr}
\end{equation}
In practice, $\delta_j$ indicates how much filter $f_j$ contributes to the classification score of the class.  
Fig.~\ref{fig:cnn:scoreDiff}{\color{blue}a} shows an example of these score differences for class \rev{\emph{car}}.
Only a few filters have high $\delta$ values, indicating they are really discriminative for the class. 
The remaining filters have low values attributable to random noise.
We consider $f_j$ to be a discriminative filter if $f_j > 2\sigma$, where $\sigma$ is the standard deviation of the distribution of $\delta$ over the 256 filters in L5.
For the car class, only 7 filters are discriminative under this definition.
Fig.~\ref{fig:cnn:scoreDiff}{\color{blue}b} shows an example of the receptive field centers of activations of the top 5 most discriminative filters, which seem to be distributed on several locations of the car.
Interestingly, on average over all classes, we find that only 9 out of 256 filters in L5 are discriminative for a particular class. The total number of discriminative filters in the network, over all 16 object classes amounts to 105. This shows that the discriminative filters are largely distributed across different object classes, with little sharing, as also observed by~\cite{agrawal14eccv}.
Hence, the network obtains its discriminative power from just a few filters specialized to each class. \\

In order to further study this, we now measure the collective impact of all discriminative filters, taken together as a set.
To do so, we generalize the discriminativeness measure defined in eq. \eqref{eq:discr} to a set of filters instead of just one.
This corresponds to simultaneously setting to zero the feature maps of all the filters in the set.
We compute it on two different sets of filters:
(1) all filters that are not discriminative and
(2) all filters that are discriminative.
The average discriminativness for (1) is 28.8, indicating that removing all non-discriminative filters has a significant impact on the class scores (for reference, the original average score using all filters is 69.2).
This is understandable given the large number of filters removed (around 247 out of 256, as there are only 9 discriminative filters per class on average). 
However, for (2), the discriminativness is much higher: 48.6. 
Therefore, we can indeed conclude that a few discriminative filters are substantially more influential than all other filters together, as the drop in class scores is greater when these filters are removed.

\begin{figure}[t]
  \begin{center}
    \includegraphics[width=\textwidth]{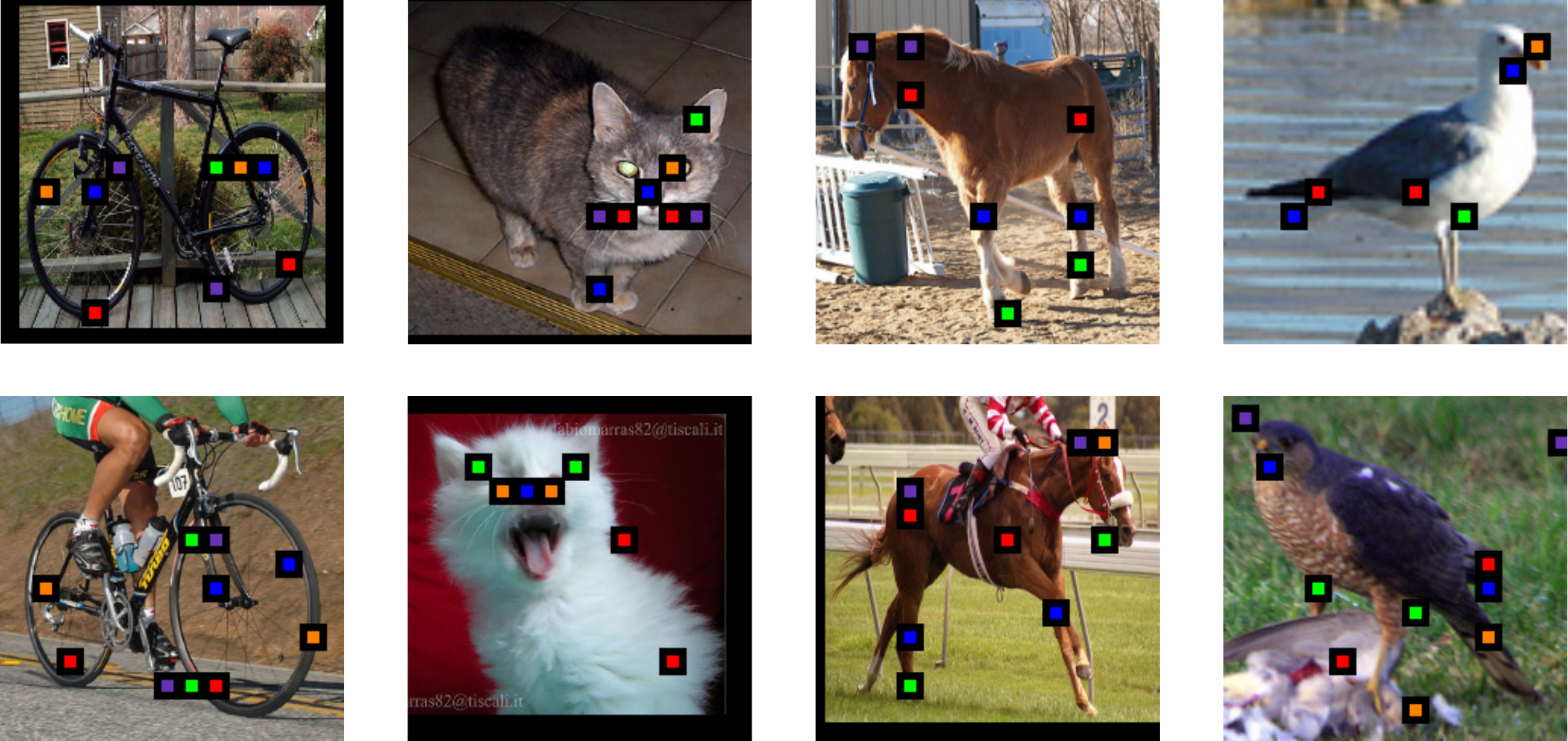}
  \end{center}
  \caption{\small Example activations of the five most discriminative filters for object class {\em bicyle, cat, horse, bird}, respectively. } 
  \label{fig:cnn:sharedFilters}
\end{figure}

Fig.~\ref{fig:cnn:sharedFilters} shows examples for other classes besides \emph{car}, where we can observe some other interesting patterns.
For example, wheels are extremely discriminative for class bicycle, in contrast to class car, where discriminative filters are more equally distributed across the whole surface of the object.
Since wheels are generally big for bicycle images, some filters specialize to subregions of the wheel, such as its bottom area.
Another interesting observation is that the discriminativeness of a semantic part might depend on the object class to which it belongs.
For example, class cat accumulates most of its most discriminative filters on parts of the head.
Interestingly, \cite{parkhi11iccv} observed a similar phenomenon with HOG features, where the most discriminative parts of cats and dogs were found to be the heads.
On the other hand, class horse tends to prefer parts of the body, such as the legs, devoting very few discriminative filters to the head. 
Besides firing on subregions of parts, some discriminative filters fire on assemblies of multiple parts or on a part with some neighboring region (e.g. the red filter for class bird is associated with both wing and tail).
\begin{figure}[t]
  \begin{center}
    \includegraphics[width=\textwidth]{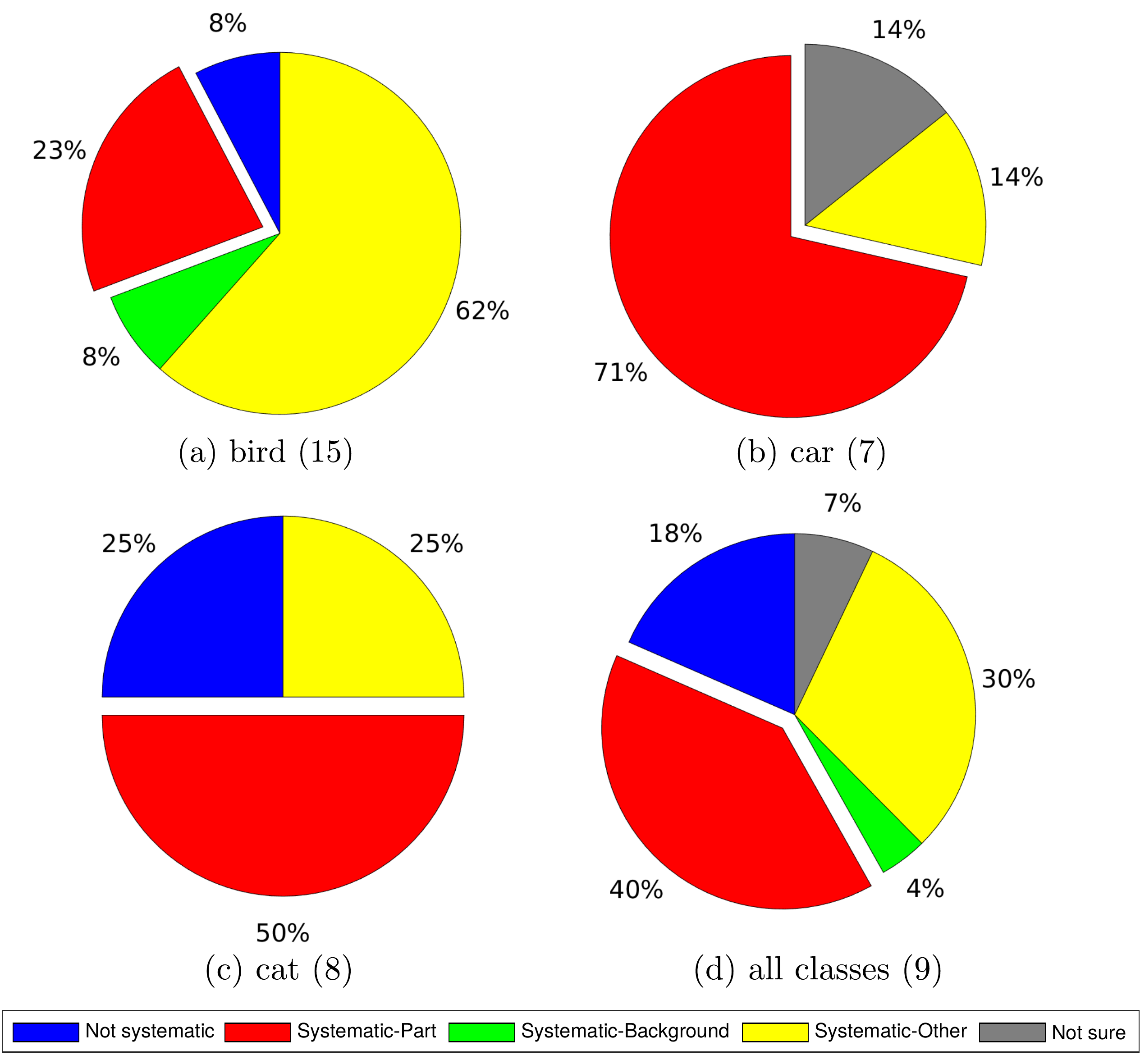}
  \end{center}
  \caption{\small Discriminative filter distributions for bird, car, cat, and for the average of all 16 classes. The number between parenthesis indicates the number of discriminative filters for each case.} 
  \label{fig:cnn:filterDiscDist}
\end{figure}

\paragraph{How many discriminative filters are also semantic?} 
We categorize now the found discriminative filters into the filter types defined in sec.~\ref{sec:cnn:humans}, using the data collected in the human experiment. 
This enables us to determine what fraction of discriminative filters are also semantic, which in turns reveals whether semantic parts are important for recognition.
Moreover, as we have defined filter types that go beyond semantic parts, we can obtain a complete list of the filter stimuli that give the network its discriminative power.

Figure~\ref{fig:cnn:filterDiscDist} shows the distribution of discriminative filters over our filter types for three object classes, and the average for all object classes.
On average, 40\% of the discriminative filters are also semantic, which translates in about 4 out of the 9 filters that are discriminative for each object class, on average.
This is a very high fraction, considering that we found only 7\% of all filters to be semantic (fig~\ref{fig:cnn:filterDist}).
This clearly indicates that the network is using semantic parts as powerful discriminative cues for recognizing object classes.
Additionally, about 4\% of the filters systematically respond to background patches, and another 30\% of the filters systematically respond to some other concept (mostly subregions or assemblies of parts).
Finally, 18\% of the filters do not correspond to any concept, a massive drop compared to the 78\% statistics over all filters (fig~\ref{fig:cnn:filterDist}). 
This distribution confirms our intuition drawn through visual inspection (fig.~\ref{fig:cnn:sharedFilters}): filters that discriminate for the network are often stimulated by some semantic part, but also by other discriminative patches such as subregions of parts.

\section{Discriminativeness of semantic parts for object recognition}
\label{sec:cnn:discrim_parts}

In the previous section we investigated how discriminative each filter is for each object class. In this section, instead, we investigate the discriminativeness of semantic parts. We look at how much each part contributes to the classification score of its object class.
We measure discriminativeness as in sec.~\ref{sec:cnn:discr_filters}, but instead of ignoring a specific filter, we now ignore a semantic part. We use the same formulation of eq.~\eqref{eq:discr}, but with different meanings for $j$, $I$ and $s_i^j$. Given a semantic part $j$, $I$ now indicates all images containing $j$, and $s_i^j$ is the score given by the network to image $I_i$ with part $j$ blacked out. 

\begin{figure}[t]
  \begin{center}
    \includegraphics[width=\textwidth]{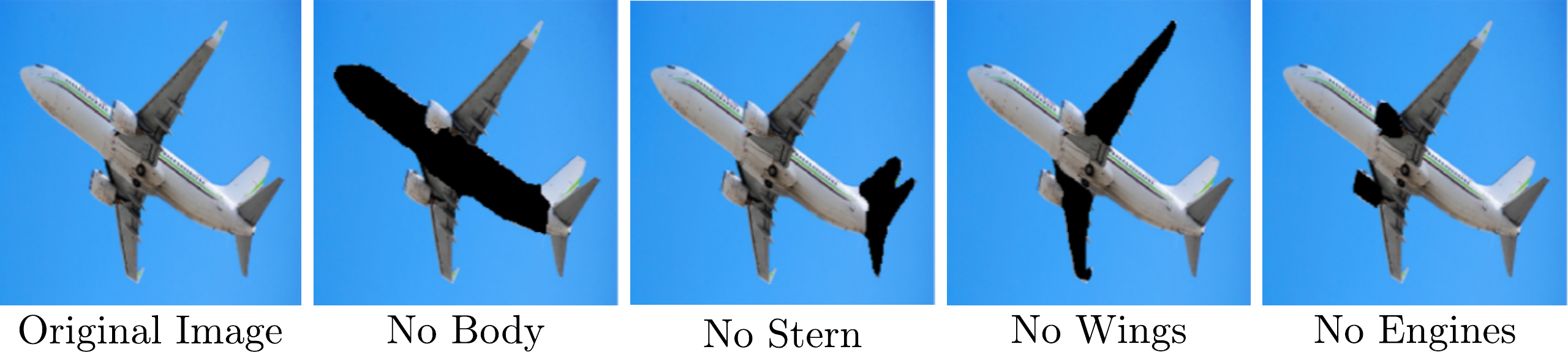}
  \end{center}
  \caption{\small Examples of input images for the network. The left-most shows the original airplane, while the others shows the same airplane, but with parts blacked out. }
  \label{fig:cnn:blackoutParts}
\end{figure}

We use the segmentation masks available in \rev{PASCAL-Part} to set to zero all the pixels of a part $j$ in each input image $I_i$, after pre-processing by subtracting the image mean (sec.~\ref{sec:cnn:exp_set}).
In this way, part $j$ is ignored and does not contribute to the classification score of the object, as all convolutional filters output 0 on blacked out regions.
The network can only rely on information from the rest of the image. If it is no longer confident about the prediction of the object class, it means that the blacked out part is discriminative for it.
We note that even if the part is blacked out, its boundary remains accessible to the network and can contribute some discriminative information.

\begin{figure*}[t]
  \begin{center}
    \includegraphics[width=\textwidth]{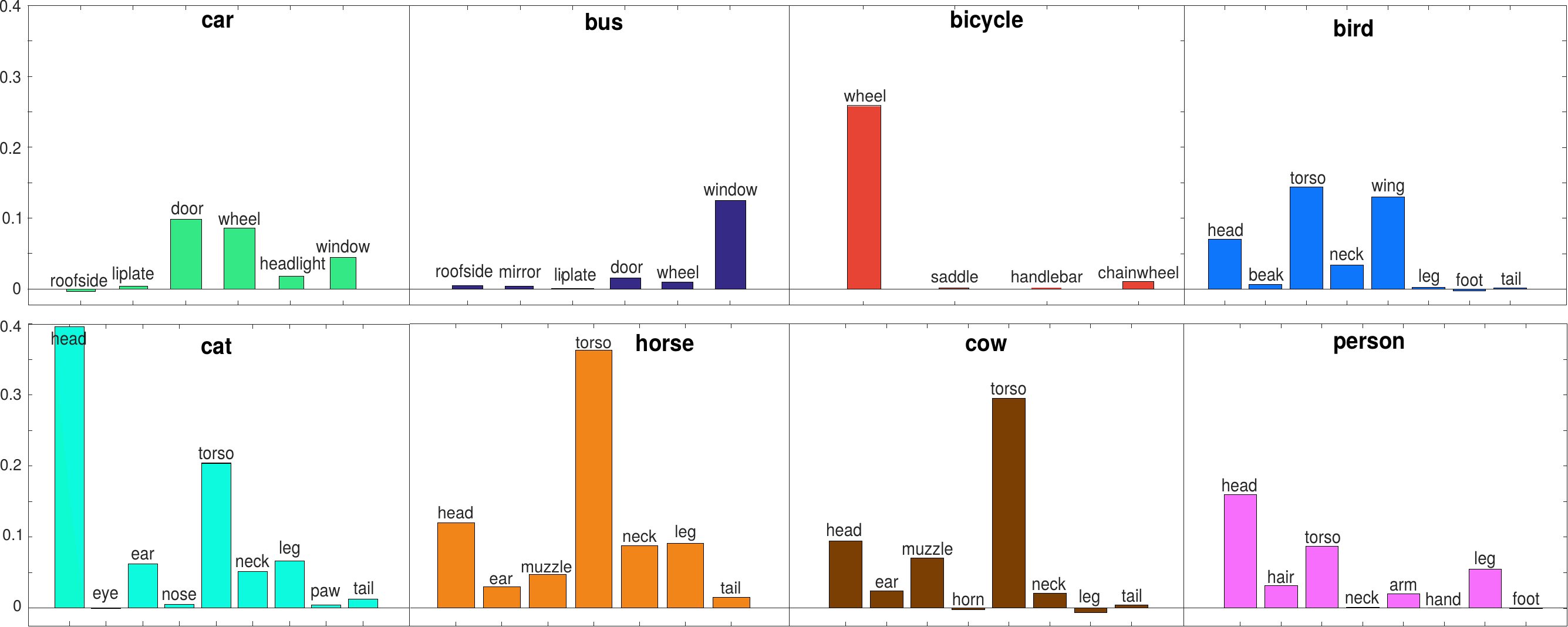}
  \end{center}
  \caption{\small Discriminativeness of \rev{PASCAL-Part} for the classification of their objects. The vertical axis indicates the difference $\delta$ between the classification score for the original image, and the score for the image after blacking out the part. We report averages over all images in an object class (higher values mean more discriminative).} 
  \label{fig:cnn:discr_parts}
\end{figure*}

We evaluate the 105 semantic parts of PASCAL-Part (sec.~\ref{sec:cnn:exp_set}). Fig.~\ref{fig:cnn:discr_parts} shows results for some examples parts of 9 object classes.
Interestingly, similar classes do not necessarily have similar discriminative parts. 
For example, the most discriminative parts for class car are \emph{door} and \emph{wheel}. But these are not very important for class bus, which is largely discriminated by the part \emph{window}. 
Moreover, \emph{torso} is very discriminative for some animals (e.g. horse and cow), but less so for others (e.g. cat and dog).
We offer two explanations for this phenomenon. First, the network seems to consider discriminative parts that are clearly visible across many instances of the object class. For example, the wheels of a car are often visible in PASCAL images, while the wheels of a bus are often occluded in the PASCAL dataset. Similarly, many images of pet animals (e.g. cat and dog) are biased towards close-ups (often occluding the body), while images of other animals typically show the whole body (e.g. horse, cow, bird).
Second, the network seems to consider more discriminative parts that have lower intra-class variation. For example, the torso is very similar across all horses, while the torso of a dog varies considerably depending on breed and size.

\begin{figure}
  \begin{center}
    \includegraphics[width=\textwidth]{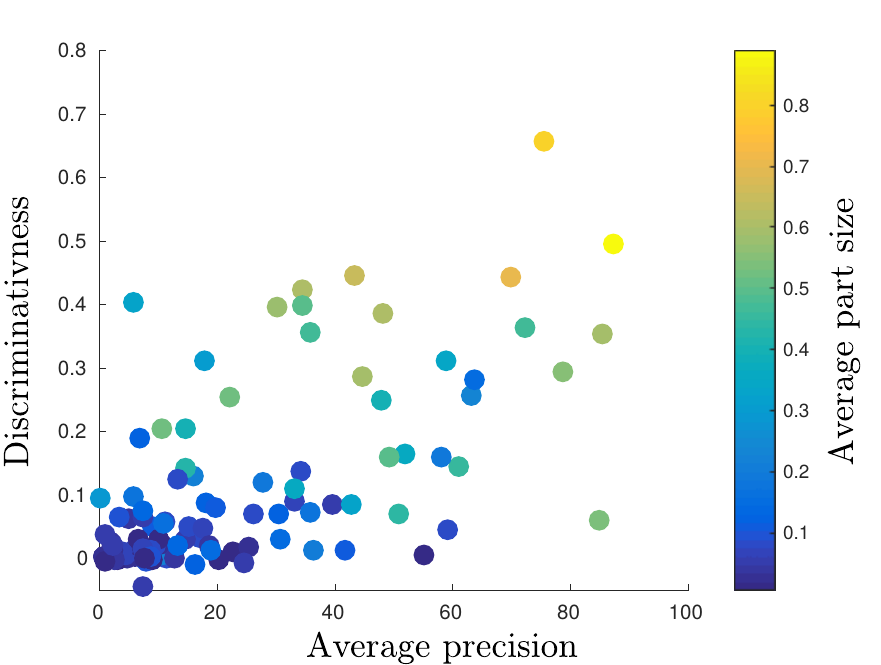}
\end{center}
  \caption{\small Correlation between the discriminativeness of a semantic part, the average precision results of sec.~\ref{sec:cnn:resAP} (mAP, GA, layer 5, table~\ref{table:part_obj}) and the average size of a part. The latter is normalized by the average size of its object. Each point corresponds to a different part. Interestingly, these measures are all correlated. } 
\label{fig:cnn:APvsPartSizevsDiscr}
\end{figure}

We also observe a strong relation between these quantitative results in fig.~\ref{fig:cnn:discr_parts} and the visual results of fig.~\ref{fig:cnn:sharedFilters}.
The top most discriminative filters activate on the most discriminative parts. For example, the only discriminative semantic part for bicycle is \emph{wheel}, which is exactly where all the activations of the most discriminative filters are. Analogously, \emph{head} is very discriminative for the class cat, \emph{wing} is discriminative for bird and \emph{leg} is somehow discriminative for \emph{horse}.

\paragraph{Correlation to average precision and part size.}
In this paragraph we look at the correlation between the discriminativeness of a semantic part, the average size of a part and the detection performance results of sec.~\ref{sec:cnn:resAP} (AP in table~\ref{table:part_obj}, GA, layer 5). Results are shown in fig.~\ref{fig:cnn:APvsPartSizevsDiscr}. 
Two interesting facts emerge. First, discriminativeness tends to increase with the average size of a part (very high PPMCC of 0.87) and second, discriminativeness correlates with how much parts emerge in the CNNs according to AP detection performance (PPMCC of 0.65). 
These are important correlations that support our analysis of sec.~\ref{sec:cnn:pascalParts}, where we observed that CNNs learn only a few of the semantic parts in their internal representation. 

Finally, note how future works that use filters as intermediate part representations (as in \cite{simon14accv,gkioxari15iccv,simon15iccv,xiao15cvpr,oquab15cvpr}) will now be able to exploit these findings to create better models for recognition.

\section{Conclusions}
\label{sec:cnn:conclusions}
We have analyzed the emergence of semantic parts in CNNs.
We have investigated whether the network's filters learn to respond to semantic parts.
In order to do so, we have associated filter stimuli to semantic parts, using two different quantitative evaluations.
In the first one, we have used ground-truth part bounding-boxes to determine how many parts emerge in the CNN for different layers, network architectures and supervision levels.
Despite promoting this emergence by providing favorable settings and multiple assists, we found that only 34 out of 105 semantic parts in PASCAL-Part dataset~\citep{chen14cvpr} emerge in AlexNet~\citep{krizhevsky12nips} finetuned for object detection~\citep{girshick14cvpr}.
\rev{This result complements previous works~\citep{zeiler14eccv,simonyan14iclr_w} by providing a more accurate, quantitative assessment and shows how the network learns to associate filters only to some part classes.} 
In the second one, we study how many filters systematically respond to semantic parts for each object class.
We found that, on average, 7\% of the filters respond to semantic parts, whereas 13\% systematically respond to other concepts, such as subregions of parts or background patches.
\rev{This filter characterization provides a more precise understanding of the internal representations learned by CNN architectures.}
Finally, we have studied how discriminative network filters and semantic parts are for the task of object recognition. 
\rev{The overlap between discriminative and semantic filters adds further insights into claims made by works based on qualitative inspection~\citep{zeiler14eccv,simonyan14iclr_w}.}


\bibliographystyle{spbasic}      
\bibliography{../../../bibtex/shortstrings,../../../bibtex/vggroup,../../../bibtex/calvin}
\end{document}